\documentclass{ecai}

\usepackage{tikz}
\usepackage{pgfplots}
\usepackage{nameref}
\usetikzlibrary{arrows,automata,shapes,trees}

\usepackage{stmaryrd}
\usepackage{algorithm}
\usepackage[noend]{algorithmic}
\usepackage{comment}
\usepackage{graphicx} 
\usepackage{pgf}
\usepackage{tikz}
\usepackage{listings}
\usepackage{amsmath}
\usepackage{amssymb}
\usepackage{proof}

\newcommand{\lft}{\hspace{-2mm}}

\newcommand{\M}[1]{\mathit{#1}}

\tikzset{global scale/.style={
    scale=#1,
    every node/.style={scale=#1}
  }
}
\newenvironment{myproof}
	{\noindent 
	{\sc Proof.}}
	{ \ \hfill $\square$\\}

\usepackage[lofdepth,lotdepth]{subfig}

\newsavebox{\tablebox}

\definecolor{light-gray}{gray}{0.5}

\makeatletter
\newcommand{\thickhline}{%
    \noalign {\ifnum 0=`}\fi \hrule height 1.5pt
    \futurelet \reserved@a \@xhline
}
\newcommand{\defterm}[1]{\underline{\textit{#1}}}

\newcommand{\GoToWorkFridays}{\namesmath{goToWorkFridays}}
\newcommand{\GoToWorkFridaysPlan}{\namesmath{goToWorkFridaysPlan}}
\newcommand{\TravelToWork}{\namesmath{travelToWork}}
\newcommand{\HaveDrinks}{\namesmath{haveDrinks}}
\newcommand{\TravelHome}{\namesmath{travelHome}}
\newcommand{\TravelHomeByBusPlan}{\namesmath{travelHmByBusPlan}}
\newcommand{\TravelHomeByTaxiPlan}{\namesmath{travelHmByTaxiPlan}}

\newcommand{\HaveCar}{\namesmath{haveCar}}
\newcommand{\FuelUsed}{\namesmath{fuelUsed}}
\newcommand{\HadDrinks}{\namesmath{intox}}
\newcommand{\DriveHomePlan}{\namesmath{driveHmPlan}}
\newcommand{\DoWork}{\namesmath{work}}

\newcommand{\mgu}{\mbox{{\sf mgu}}}

\newtheorem{lemma}[theorem]{Lemma}

\newtheorem{definitionAux}{Definition} 
\newenvironment{definition}[1][Proof]
 	{\begin{definitionAux}
 	 \begin{trivlist}
 		\item[\hskip \labelsep {\sc (#1)}]
 		\normalfont
 	}
 	{\hfill $\blacksquare$
 	 \end{trivlist}
 	 \end{definitionAux}}

\renewcommand{\vec}[1]{\mathbf{#1}}

\newcommand{\namesmath}[1]{\mbox{\textit{#1}}}
\newcommand{\propername}[1]{\text{#1}}

\newcommand{\AgentSpeak}{\propername{AgentSpeak}}

\newcommand{\CAN}{\propername{CAN}}
\newcommand{\HTN}{HTN}
\newcommand{\BDI}{BDI}

\newcommand{\CANPLAN}{\propername{CANPlan}}

\newcommand{\Navigate}{\namesmath{nav}}
\newcommand{\Move}{\namesmath{move}}

\newcommand{\ObtainSoilResults}{\namesmath{getSoilRes}}
\newcommand{\TransmitSoilResults}{\namesmath{transmitRes}}

\newcommand{\PickSoilSample}{\namesmath{pickSoil}}

\newcommand{\AnalyseSoilSample}{\namesmath{analyseSoil}}

\newcommand{\GetMoistureContent}{\namesmath{getMoisture}}

\newcommand{\GetSoilParticleSize}{\namesmath{getSoilSize}}

\newcommand{\DropSoilSample}{\namesmath{dropSoil}}
\newcommand{\EstablishConnection}{\namesmath{establishCon}}
\newcommand{\UploadResults}{\namesmath{uploadRes}}
\newcommand{\SendResults}{\namesmath{sendRes}}

\newcommand{\BreakConnection}{\namesmath{breakCon}}
\newcommand{\CalibrateViaGPS}{\namesmath{calib}}

\newcommand{\PerformSoilExperiment}{\namesmath{doSoilExp}}

\newcommand{\false}{\namesmath{false}}
\newcommand{\true}{\namesmath{true}}
\newcommand{\addedSignature}{\namesmath{addedSignature}}
\newcommand{\sendEmail}{\namesmath{sendMail}}
\newcommand{\sent}{\namesmath{sent}}

\newcommand{\ExploreSoilLocation}{\namesmath{explore}}
\newcommand{\ResultsTransmitted}{\namesmath{resultsTransmitted}}
\newcommand{\HaveMoistureContent}{\namesmath{haveMoistureContent}}
\newcommand{\HaveParticleSize}{\namesmath{haveParticleSize}}

\newcommand{\HMC}{\namesmath{hMC}}
\newcommand{\HPS}{\namesmath{hPS}}
\newcommand{\IN}{\namesmath{in}}
\newcommand{\HSS}{\namesmath{hSS}}

\newcommand{\Calibrated}{\namesmath{calibrated}}

\newcommand{\LAt}{\namesmath{at}}

\newcommand{\RT}{\namesmath{rT}}

\newcommand{\CE}{\namesmath{cE}}
\newcommand{\CCA}{\namesmath{cal}}

\newcommand{\HaveSoilSample}{\namesmath{haveSoilSample}}
\newcommand{\ConnectionEstablished}{\namesmath{connectionEstablished}}

\newcommand{\CA}{\mbox{$\cal A$}}

\newcommand{\BB}{\mbox{$\cal B$}}

\newcommand{\RR}{\mbox{$\cal R$}}

\newcommand{\tuple}[1]{\M{\langle #1 \rangle}}
\newcommand{\config}[1]{\M{\langle #1 \rangle}}
\newcommand\plane[1]{\planeaux!#1!}
\newcommand\planev[1]{\planeVaux!#1!}
\def\planeaux!#1:#2<-#3!{\M{#1 \mbox{\rm:} #2\; \leftarrow #3}}
\def\planeVaux!#1[#2]:#3<-#4!{#1(\vec{#2}) \mbox{\rm:} #3\; \leftarrow #4}

\def\planeaux!#1<-#2!{\M{#1 \leftarrow #2}}

\newcommand{\transition}{\longrightarrow}       % regular transition
\newcommand{\transitionlabel}[1]{\M{\mathsf{#1}}}
\newcommand{\bdi}{bdi}
\newcommand{\htn}{plan}

\newcommand{\labelhtn}{plan}

\newcommand{\transitiontype}[1]{\overset{\transitionlabel{#1}}{\transition}}

\newcommand{\transitionhtn}{\overset{\transitionlabel{\labelhtn}}{\transition}}
\newcommand{\transitionhtnstar}
{\overset{\transitionlabel{\labelhtn}_{*}}{\transition}}

\newcommand{\Summarise}{\mbox{\textsf{Summ}}}
\newcommand{\SummarisePlanBody}{\mbox{\textsf{SummPlan}}}
\newcommand{\SummariseEvent}{\mbox{\textsf{SummEvent}}}

\newcommand{\Plan}{\mbox{\sf Plan}}

\newcommand{\pnil}{\mbox{\textit{nil}}}

\newcommand{\pguardaltl}[1]{\mbox{$\altl #1 \altr$}} % guarded alternatives
\newcommand{\altl}{\llparenthesis}
\newcommand{\altr}{\rrparenthesis}

\begin{document}

\title{Addendum to: Summary Information for Reasoning About Hierarchical Plans}

\author{Lavindra de Silva\institute{Institute for Advanced Manufacturing, University of Nottingham, Nottingham, UK, e-mail: lavindra.desilva@nottingham.ac.uk} \and Sebastian Sardina\institute{RMIT University, Melbourne, Australia, e-mail: \{ssardina, linpa\}@cs.rmit.edu.au} \and Lin Padgham$^2$}

\maketitle

%%%%%%%%%%%%%%%%%%%%%%%%%%%%%%%%%%%%%%%%%%%%%%%%%%%%%%%%%%%%%%%%%%%%%%%%%%%%%%%
%%%%%%%%%%%%%%%%%%%%%%%%%%%%%%%%%%%%%%%%%%%%%%%%%%%%%%%%%%%%%%%%%%%%%%%%%%%%%%%
%%%%%%%%%%%%%%%%%%%%%%%%%%%%%%%%%%%%%%%%%%%%%%%%%%%%%%%%%%%%%%%%%%%%%%%%%%%%%%%

\begin{abstract}

Hierarchically structured agent plans are 
important for efficient planning and acting, 
and they also serve (among other things) 
to produce ``richer'' 
classical plans, composed not just of a 
sequence of primitive actions, but also 
``abstract'' ones representing the
supplied hierarchies. A crucial step 
for this and other approaches is deriving
precondition and effect ``summaries'' from 
a given plan hierarchy. This paper provides 
mechanisms to do this for more pragmatic 
and conventional
hierarchies than in the past. To this end,
we formally define the notion of a precondition 
and an effect for a hierarchical plan; we 
present data structures and algorithms 
for automatically deriving this information; 
and we analyse the properties of the presented 
algorithms. We conclude the paper by detailing 
how our algorithms may be used together with 
a classical planner in order to obtain abstract 
plans.
 
\end{abstract}

%%%%%%%%%%%%%%%%%%%%%%%%%%%%%%%%%%%%%%%%%%%%%%%%%%%%%%%%%%%%%%%%%%%%%%%%%%%%%%%
%%%%%%%%%%%%%%%%%%%%%%%%%%%%%%%%%%%%%%%%%%%%%%%%%%%%%%%%%%%%%%%%%%%%%%%%%%%%%%%
%%%%%%%%%%%%%%%%%%%%%%%%%%%%%%%%%%%%%%%%%%%%%%%%%%%%%%%%%%%%%%%%%%%%%%%%%%%%%%%
\section{INTRODUCTION}\label{sec:create-ops}
%%%%%%%%%%%%%%%%%%%%%%%%%%%%%%%%%%%%%%%%%%%%%%%%%%%%%%%%%%%%%%%%%%%%%%%%%%%%%%%

This paper provides effective techniques for automatically extracting abstract actions and plans from a supplied hierarchical agent plan-library. 
Hierarchically structured agent plans are appealing for efficient acting and planning as they embody an expert's domain control knowledge, which can greatly speed up the reasoning process and cater for non-functional requirements. 
Two popular approaches based on such representations are Hierarchical Task Network (HTN)~\cite{erol94htn,GhallabNT:04-Planning} planning and Belief-Desire-Intention (BDI) ~\cite{rao:95} agent-oriented programming.
While HTN planners ``look ahead'' over a supplied  collection of hierarchical plans to determine whether a task has a viable decomposition, BDI agents interleave this process with acting in the real world, thereby trading off solution quality for faster responsiveness to environmental  changes. Despite these differences, HTN and BDI systems are closely related in both syntax and semantics, making it possible to translate between the two representations~\cite{sardina:aamas:06,sardina:jaamas}. 

While HTN planning and BDI execution are concerned with decomposing hierarchical structures (offline and online, respectively), one may perform other kinds of reasoning with them that do not necessarily require or amount to decompositions.
%
%Besides their most common decomposition type of use in HTN planning and BDI execution, hierarchical structures can also support direct reasoning at the abstract level.
%%
For example, \cite{clement:jair:07} and \cite{john-ij,John-03} perform reasoning to coordinate the execution of abstract steps, so as to preempt potential negative interactions or exploit positive ones.  
In~\cite{desilva:aamas:09}, the authors propose a novel application of such hierarchies to produce ``richer'' classical plans composed not just of sequences of primitive actions, but also of ``abstract'' steps. Such abstract plans are particularly appealing in the context of BDI and HTN systems because they respect and re-use the domain control knowledge inherent in such systems, and they provide flexibility and robustness: if a refinement of one abstract step happens to fail, another option may be tried to achieve the step.

A pre-requisite for these kinds of reasoning is the availability of meaningful preconditions and effects for abstract steps (i.e., compound tasks in HTN systems or event-goals in BDI languages).
Generally, this information is not supplied explicitly, but embedded within the decompositions of an abstract step.
%This information is generally only implicitly provided through the possible decompositions.
%%
This paper provides techniques for extracting this information automatically. 
The algorithms we develop are built upon those of~\cite{clement:jair:07} and~\cite{john-ij,John-03}, which calculate offline the precondition and effect ``summaries'' of HTN-like hierarchical structures that define the agents in a multi-agent system, and use these summaries at runtime to coordinate the agents or their plans.
The most important difference between these existing techniques and ours is that the former are framed in a propositional language, whereas ours allow for first-order variables. This is fundamental when it comes to practical applicability, as any realistic BDI agent program will make use of variables.
A nuance worth mentioning between our work and that of Clement et al. is that the preconditions we synthesise are standard classical precondition formulas (with disjunction), whereas their preconditions are (essentially) two sets of literals: the ones that must hold at the start of \textit{any} successful execution of the entity, and the ones that must hold at the start of \textit{at least one} such execution. Yao et al. \cite{yao:16} extend the above two strands of work to allow for concurrent steps within an agent's plan, though still not first-order variables.

Perhaps the only work that computes summaries 
(``external conditions'') of hierarchies specifying 
first-order variables is \cite{tsuneto:98}. 
The authors automatically extract a weaker form of 
summary information (what we call ``mentioned'' 
literals) to inform the task selection 
strategy of the UMCP HTN planner: 
tasks that can possibly establish or threaten the 
applicability of other tasks are explored first. 
They show that even weak summary information 
can significantly reduce backtracking and
increase planning speed. However, the authors 
only provide insights into their algorithms for 
computing summaries.

We note that we are only concerned here with how to extract abstract actions (with corresponding preconditions and effects), and eventually   
 abstract plans, from a hierarchical know-how structure. Consequently, unlike existing useful and interesting work~\cite{rao:aaai:98,sheila:kr:08,sheila:icaps:07,wayne:09}, our approach does not directly involve \emph{guiding a planner} toward finding a suitable
primitive plan. We also do not aim to build new ``macro'' actions from sample primitive solution plans, as done in~\cite{macroff}, for example. 

% Finding a first abstract plan 
% is also a prerequisite in \cite{desilva:aamas:09}, who 
% improve it to extract its ``maximally abstract'' and 
% non-redundant subset. While these authors do provide 
% some insights into their definitions and algorithms, 
% they do not, however, give details about how abstract 
% actions and plans are obtained.
%  
% In both \cite{desilva:aamas:09} and our work, 
% abstract operators represent BDI event-goals 
% (HTN compound tasks). Since event-goals are not, 
% \textit{per se} attached with precondition and 
% effect information, we derive these automatically 
% from the structure of an event-goal's hierarchy, 
% combined with knowledge about the low-level, primitive 
% actions. 
% Intuitively, the precondition of 
% an event-goal encodes the conditions under which 
% it will successfully execute, 
% %
% and its postcondition defines what conditions will 
% hold regardless of the decompositions chosen to achieve 
% the event-goal (its \textit{definite effects}), as well 
% as which ones will hold in at least one trace through 
% its decompositions (its \textit{possible effects}). 
% %
% While definite effects are crucial for reasoning about abstract 
% operators for planning/coordination, possible effects are 
% also necessary to be able to compute definite effects, and 
% ascertain whether an abstract plan found or a potential 
% coordination of plans is viable.  

Thus, the contributions of this paper are as follows. First, we develop formal definitions for the notions of a precondition and an effect of an (abstract) event-goal. 
Second, we develop algorithms and data structures for deriving precondition and effect summaries from an event-goal's hierarchy. Unlike past work, we use a typical BDI agent programming language framework; in doing so, we allow for variables in agent programs---an important requirement in practical systems. Our chosen BDI agent programming language cleanly incorporates the syntax and semantics of HTN planning as a built-in feature, making our work immediately accessible to both communities.
Finally, we show how derived event-goal summaries may be used together with a classical planner in order to obtain abstract plans (which can later be further refined, if desired, to meet certain properties~\cite{desilva:aamas:09}).

%%%%%%%%%%%%%%%%%%%%%%%%%%%%%%%%%%%%%%%%%%%%%%%%%%%%%%%%%%%%%%%%
%%%%%%%%%%%%%%%%%%%%%%%%%%%%%%%%%%%%%%%%%%%%%%%%%%%%%%%%%%%%%%%%
%%%%%%%%%%%%%%%%%%%%%%%%%%%%%%%%%%%%%%%%%%%%%%%%%%%%%%%%%%%%%%%%
\section{THE HIERARCHICAL FRAMEWORK}\label{sec:prelim}
% \section{The \CANPLAN\ syntax and semantics}
%%%%%%%%%%%%%%%%%%%%%%%%%%%%%%%%%%%%%%%%%%%%%%%%%%%%%%%%%%%%%%%%

Our definitions, algorithms, and results are based on the formal machinery provided by the \CANPLAN~\cite{sardina:jaamas} language and operational semantics. 
While designed to capture the essence of BDI agent-oriented languages, it directly relates to other hierarchical representations of procedural knowledge, such as HTN planning~\cite{erol94htn,GhallabNT:04-Planning}, both in syntax and semantics.

A \CANPLAN\ \BDI\ agent is created by the specification
of a \textit{belief base} $\BB$, i.e., a set of ground atoms,
a \textit{plan-library} $\Pi$, i.e., a set of plan-rules,
and an \textit{action-library} $\Lambda$, i.e., a set of
action-rules.
A plan-rule is of the form $\planev{e[v]:\psi<-P}$, 
where $e(\vec{v})$ is an \textit{event-goal}, $\vec{v}$ 
is a vector of distinct variables, $\psi$ is the 
\textit{context condition}, and $P$ is a \textit{plan-body} 
or \textit{program}.\footnote{In \cite{sardina:jaamas} 
an event-goal is of the form $e(\vec{t})$ where 
$\vec{t}$ is a vector of terms. Here, we replace 
$\vec{t}$ with $\vec{v}$ and assume WLOG that 
$\forall t_i \in \vec{t}, \psi \supset (t_i = v_i)$,
where $v_i \in \vec{v}$.}
The latter is made up of the following components: primitive 
actions ($act$) that the agent can execute directly; 
operations to add ($+b$) and remove ($-b$) beliefs; 
tests for conditions ($?\phi$); and event-goal 
programs ($!e$), which are simply event-goals
combined with the label ``$!$''. These components 
are composed using the sequencing construct $P_1 ; P_2$. 
While the original definition of a plan-rule also 
included declarative goals and the ability to specify 
partially ordered programs \cite{Winikoff02}, we leave 
out these constructs here and focus only on an \AgentSpeak-like 
\cite{agentspeak}, typical BDI agent programming language.
 
There are also additional constructs used by \CANPLAN\ 
internally when attaching semantics to constructs. These 
are the programs $\pnil$, $P_1 \triangleright P_2$, and 
$\pguardaltl{\psi_1:P_1, \ldots, \psi_n:P_n}$.
Intuitively, $\pnil$ is the empty program, which
indicates that there is nothing left to execute; 
program $\pguardaltl{\psi_1:P_1, \ldots, \psi_n:P_n}$ 
represents the plan-rules that are relevant for 
some event-goal; and program $P_1 \triangleright P_2$ 
realises failure recovery:
program $P_1$ should be tried first, failing which 
$P_2$ should be tried.
The complete language of \CAN, then, is described by 
the grammar
\[\small
\begin{array}{lll}
P & ::=  & \pnil \mid act \mid\ ?\phi \mid {+}b \mid {-}b \mid\ 
        !e \mid  P_1;P_2 \mid P_1 \triangleright P_ 2 \mid\\ 
  &     & \pguardaltl{\psi_1:P_1, \ldots, \psi_n:P_n}. 
\end{array}
\]
 
The behaviour of a \CANPLAN\ agent is defined by a 
set of derivation rules in the style of Plotkin's
structural single-step operational semantics 
\cite{plotkin}. 
The \textit{transition relation} on a configuration
is defined using one or more derivation rules.
Derivation rules have an \textit{antecedent} and
a \textit{conclusion}: the antecedent can either
be empty, or it can have transitions and auxiliary
conditions; the conclusion is a single transition.
A \textit{transition} $C \transition C'$ within 
a rule denotes that configuration $C$ yields 
configuration $C'$ in a single execution 
step, where a configuration is the tuple 
$\config{\BB,\CA,P}$ composed of a belief 
base $\BB$, a program $P$, and the sequence
$\CA$ of actions executed so far.  
%
%A transition $\transitionstar$ denotes the 
%reflexive transitive closure of $\transition$,
Construct $C \transitiontype{t} C'$ denotes 
a transition of type $\transitionlabel{t}$, where 
$\transitionlabel{t} \in \{\transitionlabel{\bdi}, 
\transitionlabel{\htn}\}$; when no label is specified 
on a transition both types apply.  
Intuitively, $\transitionlabel{\bdi}$-type 
transitions are used 
for the standard \BDI\ execution cycle, and 
$\transitionlabel{\htn}$-type transitions for 
(internal) deliberation steps within a 
\textit{planning} context. By distinguishing 
between these two types of transitions, certain 
rules can be disallowed from being 
used in a planning context, 
such as those dealing with BDI-style failure handling.

We shall describe three of the \CANPLAN\ 
derivation rules. The rule below 
%collects 
%all relevant plan-rules for $e$, i.e., the
%ones whose handling event-goal unifies with $e$,
%and stores them in $\pguardaltl{\Delta}$;
%$\mgu$ stands for ``most general unifier'' \cite{lloyd:87}.
%Specifically, the rule states that
states that a configuration
$\config{\BB,\CA,!e}$ evolves into a
configuration $\config{\BB,\CA,\pguardaltl{\Delta}}$ 
(with no changes to $\BB$ and $\CA$)
in one $\transitionlabel{\bdi}$- or 
$\transitionlabel{\htn}$-type execution 
step, with $\pguardaltl{\Delta}$ being the
set of all relevant plan-rules for $e$, i.e., the
ones whose handling event-goal unifies with $e$;
$\mgu$ stands for ``most general unifier'' \cite{lloyd:87}.
From $\pguardaltl{\Delta}$, an applicable
plan-rule---one whose context condition
holds in $\BB$---is selected by another
derivation rule and the associated plan-body scheduled
for execution.
%\vspace{-0.5mm}  
\[
\small
\begin{array}{c}
\infer[]
{\config{\BB, \CA,!e} \transition \config{\BB, \CA,\pguardaltl{\Delta}}}
{
\Delta = \{ \psi_i\theta:P_i \theta\ |\ \plane{e':\psi_i<-P_i} \in \Pi \land
\theta=\mgu(e,e') \} }
\end{array}
\]
%\vspace{-0.5mm}  

The $\Plan$ construct incorporates HTN planning 
as a built-in feature of the semantics.
The main rule defining the construct states that a configuration
$\config{\BB,\CA,\Plan(P)}$ evolves into a
configuration $\config{\BB',\CA',\Plan(P')}$ in
one $\transitionlabel{\bdi}$-type execution step
if the following two conditions hold: \textit{(i)}
configuration $\config{\BB,\CA,P}$ yields 
configuration $\config{\BB',\CA',P'}$ in one
$\transitionlabel{\htn}$-type execution step, 
and \textit{(ii)} it is possible to reach a 
\textit{final} configuration $\config{\BB'',\CA'',\pnil}$
from $\config{\BB',\CA',P'}$ in a finite number
of $\transitionlabel{\htn}$-type execution steps.
Thus, executing the single $\transitionlabel{\bdi}$-type
step necessitates zero or more internal ``look ahead'' 
steps that check for a successful HTN execution of $P$.
 
Unlike plan-rules, any given action program 
will have exactly one associated action-rule 
in the action-library $\Lambda$. 
Like a STRIPS operator, an action-rule
$\plane{act:\psi<-\Phi^+;\Phi^-}$ is such that
$act$ is a symbol followed by a vector of distinct
variables, and all variables free in $\psi$,
$\Phi^+$ (the add list) and $\Phi^-$ (the
delete list) are also free in $act$.
We additionally expect any action-rule 
$\plane{act:\psi<-\Phi^-;\Phi^+}$ to be 
\textit{coherent}: that is, for all ground instances 
$act\theta$ of $act$, if $\psi\theta$ is consistent, then 
$\Phi^+\theta \cup \{\neg b \mid b \in \Phi^-\theta\}$ is 
consistent. For example, while the rule $R$ corresponding to
an action $move(X,Y)$ with precondition $at(X) \land \neg 
at(Y)$ (or $X \neq Y$) and postcondition $\neg at(X) \land 
at(Y)$ is coherent, the same rule with precondition $true$ 
is not, as there will then be a ground instance of $R$ 
such that its precondition is consistent but its postcondition
is not: both its add and delete lists contain the same atom.

%%%%%%%%%%%%%%%%%%%%%%%%%%%%%%%%%%%%%%%%%%%%%%%%%%%%%%%%%%%%%%%%%%%%%%%%%%%%%%%%
%%%%%%%%%%%%%%%%%%%%%%%%%%%%%%%%%%%%%%%%%%%%%%%%%%%%%%%%%%%%%%%%%%%%%%%%%%%%%%%%
%%%%%%%%%%%%%%%%%%%%%%%%%%%%%%%%%%%%%%%%%%%%%%%%%%%%%%%%%%%%%%%%%%%%%%%%%%%%%%%%

\section{ASSUMPTIONS}
\label{sec:assum}

We shall now introduce some of the definitions used in the 
rest of the paper and concretise the rest of our assumptions. 
As usual, we use $\vec{x}$ and $\vec{y}$ to denote vectors 
of distinct variables, and $\vec{t}$ to denote a vector of 
(not necessarily distinct) terms. Moreover, since the 
language of \CANPLAN\ allows variables in programs, we 
shall frequently make use of the notion of a 
\textit{substitution} \cite{lloyd:87}, which is
a finite set $\{x_1/t_1,\ldots,x_n/t_n\}$ of elements where $x_1,\ldots,x_n$ are distinct variables and each $t_i$ is a term with $t_i \neq x_i$. We use $E\theta$ to denote the expression obtained from  any expression $E$ by simultaneously replacing each
occurrence of $x_i$ in $E$ with $t_i$, for all $i \in \{1,\ldots,n\}$. 
 
We assume that the plan-library does 
not have recursion. Formally, we assume that a
\textit{ranking} exists for the plan-library, 
i.e., that it is always possible to give a child 
a smaller rank (number) than its parent. We
define a ranking as follows.

\begin{definition}[Ranking]\label{def:ranking}
A \defterm{ranking} for a plan-library $\Pi$ is a 
function $\RR_\Pi : E_{\Pi} \mapsto \mathbb{N}_0$ 
from event-goal types mentioned in $\Pi$ to natural 
numbers, such that for all event-goals 
$e_1, e_2 \in E_{\Pi}$ where $e_2$ is the same type 
as some $e_3 \in children(e_1,\Pi)$, we have that
$\RR_\Pi(e_1) > \RR_\Pi(e_2)$.\footnotemark 
\end{definition}
\footnotetext{We define the function $children(\hat{e},\Pi)=
\{e \mid \plane{e':\psi<-P} \in \Pi,$
$\hat{e} \text{ and } e' \text{ are the same type, } 
P \text{ mentions } !e\}$, where two event-goals are
\textit{the same type} if they have the same predicate
symbol and arity. 
The \textit{type} of an event-goal $e(\vec{t})$
is defined as $e(\vec{x})$, where
$|\vec{x}|=|\vec{t}|$.}

In addition, we define the following two related notions: 
first, given an event-goal type $e$, $\RR_\Pi(e)$ denotes 
the \textit{rank} of $e$ in $\Pi$; and second, given any 
event-goal $e(\vec{t})$ mentioned in $\Pi$, we define 
$\RR_\Pi(e(\vec{t})) = \RR_\Pi(e(\vec{x}))$ (where 
$|\vec{x}|=|\vec{t}|$), i.e., the rank of an event-goal 
is equivalent to the rank of its type. In order that
these and other definitions also apply to event-goal 
\textit{programs}, we sometimes blur the distinction 
between event-goals $e$ and event-goal programs $!e$.

Finally, we assume that context conditions are
written with appropriate care. Specifically, if there
is no environmental interference, whenever a plan-rule
is applicable it should be possible to successfully
execute the associated plan-body without any failure
and recovery; this disallows rules such as
$\plane{e:\true<-?\false}$.
Our definition makes use of the notion of a \textit{projection}: 
given any configuration $\config{\BB,\CA,P}$, we define 
the \defterm{projection} of the first component of the tuple as 
$C|_{\BB}$, the second as $C|_{\CA}$, and the third as $C|_{P}$.

\begin{definition}[Coherent Library]\label{def:safeLib}
A plan-library $\Pi$ is \defterm{coherent} if for all 
rules $\plane{e:\psi<-P} \in \Pi$, ground instances 
$e\theta$ of $e$, and belief bases $\BB$, whenever $\BB \models 
\psi\theta\theta'$ (where $\psi\theta\theta'$ is ground)
there is a successful \HTN\ execution $C_1 \cdot\ldots\cdot 
C_n$ of $P\theta\theta'$ (relative to $\Pi$) with 
$C_1|_{\BB} = \BB$. A \defterm{successful \HTN execution} 
of a program $P$ relative to a plan-library is any finite 
sequence of configurations $C_1 \cdot\ldots\cdot C_n$ such 
that $C_1|_{P} = P$, $C_n|_{P} = \pnil$, and for all 
$0 < i < n , C_i \transitionhtn C_{i+1}$.
\end{definition}

Intuitively, the term \textit{HTN execution} simply denotes 
a BDI execution in which certain BDI-specific derivation 
rules associated with failure and recovery have not been 
used.

%%%%%%%%%%%%%%%%%%%%%%%%%%%%%%%%%%%%%%%%%%%%%%%%%%%%%%%%%%%%%%%%%%%%%%%%%%%%%%%%
%%%%%%%%%%%%%%%%%%%%%%%%%%%%%%%%%%%%%%%%%%%%%%%%%%%%%%%%%%%%%%%%%%%%%%%%%%%%%%%%
%%%%%%%%%%%%%%%%%%%%%%%%%%%%%%%%%%%%%%%%%%%%%%%%%%%%%%%%%%%%%%%%%%%%%%%%%%%%%%%%

\section{SUMMARY INFORMATION}
\label{sec:formalise}

We can now start to define what we mean by preconditions 
and postconditions/effects of event-goals; some of these definitions are 
also used later in the algorithms. As a first step we 
define these notions for our most basic programs. 

A basic program is either an \textit{atomic program} 
or a \textit{primitive program}. Formally, 
a program $P$ is an \defterm{atomic program} (or
simply \textit{atomic}) if 
$P = !e \mid act \mid +b \mid -b \mid ?\phi$, and 
$P$
is a \defterm{primitive program} if $P$ is an atomic 
program that is not an event-goal program.
Then, like the postcondition of a STRIPS action, the 
\textit{postcondition} of a primitive program is simply 
the atoms that will be added to and removed from the 
belief base upon executing the program. Formally, 
the \defterm{postcondition} of a primitive program $P$ 
relative to an action-library $\Lambda$, denoted 
$post(P, \Lambda)$, is the set of literals 
$post(P,\Lambda) =$
\[\small \left\{
\begin{array}{ll}\label{def:postcond}
         \emptyset 	& \mbox{if $P = ?\phi$},\\
         \{b\} 		& \mbox{if $P = +b$},\\
         \{\neg b\} 	& \mbox{if $P = -b$},\\
         \Phi^+\theta \cup \{\neg b \mid b \in \Phi^-\theta\} 
			& \mbox{if $P = act$ and there exists} \\
			& \mbox{an $\plane{act':\psi<-\Phi^+;\Phi^-} \in \Lambda$}\\
			& \mbox{such that $act = act'\theta$}.
\end{array} \right. \]\noindent
The postcondition of a test condition is the empty
set because executing a test condition does not result 
in an update to the belief base. The postcondition of 
an action program 
is the combination of the add list and delete list 
of the associated action-rule, after applying the 
appropriate substitution. 

While this notion of a postcondition as applied to 
a primitive program is necessary for our algorithms 
later, we do not also need the matching notion of
a \textit{precondition} of a primitive program. 
Such preconditions are already accounted for in 
context conditions of plan-rules, by virtue 
of our assumption (Definition \ref{def:safeLib})
that the latter are coherent. 
What we do require, however, is the notion of
a \textit{precondition} as applied to an event-goal. 
This is defined as any formula such that whenever 
it holds in some state there is at least one 
successful \HTN\ execution of the event-goal 
from that state. 

\begin{definition}[Precondition]\label{def:summ-prec}
A formula $\phi$ is said to be a \defterm{precondition} of an event-goal 
$!e$ (relative to a plan- and an action-library) if for all 
ground instances $!e\theta$ of $!e$ and belief bases $\BB$, 
whenever $\BB \models \phi\theta$, there exists a 
successful \HTN\ execution $C_1 \cdot\ldots\cdot C_n$ of 
$!e\theta$, where $C_1|_{\BB} = \BB$.
\end{definition}

Unlike the postcondition of a primitive program, the postcondition 
of an event-goal program---and indeed any arbitrary program $P$---is 
non-deterministic: it depends on what plan-rules
are chosen to decompose $P$. There are, nonetheless, certain 
effects that will be brought about irrespective of such choices. 
We call these \textit{must literals}: literals that hold at the end 
of \textit{every} successful \HTN\ execution of $P$. 

\begin{definition}[Must Literal]\label{def:must-summ}
Let $P$ be a program and $l$ a literal where its variables are 
free in $P$. Then, $l$ is a \defterm{must literal} of $P$ (relative 
to a plan- and action-library) if for any ground instance $P\theta$ 
of $P$ and successful \HTN\ execution $C_1 \cdot\ldots\cdot C_n$ 
of $P\theta$, we have that $C_n|_{\BB} \models l\theta$.\end{definition}

A desirable consequence of the two definitions above is 
that any given set of must literals of an event-goal, like 
the postcondition of an action, is consistent whenever the
event-goal's precondition is consistent. 

\begin{theorem}\label{thm:summ-alg-consistent}
Let $e$ be an event-goal, $\phi$ a precondition of $e$ 
(relative to a plan-library $\Pi$ and an action-library $\Lambda$), 
and $L^{mt}$ a set of must literals of $e$ (relative to $\Pi$ and 
$\Lambda$). Then, for all ground instances $e\theta$ of $e$, if 
$\phi\theta\theta'$ is consistent for some ground substitution 
$\theta'$, then so is $L^{mt}\theta$.
\end{theorem}
\ifdefined\withsketch
\begin{myproof}
We prove this by contradiction.
First, note that since $L^{mt}\theta$ is a set of ground literals (by
Definition \ref{def:must-summ} and because $e\theta$ is ground), if
$L^{mt}\theta$ is consistent, then for all literals $l,l' \in L^{mt}$
it is the case that $l\theta \neq \overline{l'}\theta$.\footnote{The 
\textit{complement} of a literal $l \in \{a, \neg a\}$, denoted by
$\overline{l}$, 
is $a$ if $l = \neg a$, and $\neg a$ otherwise.} If we assume
that the theorem does not hold, then there must be a ground instance
$e\theta$ of $e$ such that $\phi\theta\theta'$ is consistent for some
ground substitution $\theta'$, but $l\theta = \overline{l'}\theta$
for some $l,l' \in L^{mt}$.

Since $\phi\theta\theta'$ is consistent, it is not difficult to show
there is a belief base $\BB$ such that $\BB \models \phi\theta\theta'$.
Then, by Definition \ref{def:summ-prec} there is also a successful HTN
execution $C_1 \cdot\ldots\cdot C_n$ of $e\theta$ with $C_1|_{\BB} = \BB$.
Moreover, since $l,l'$ are must literals of $e$ and $e\theta$ is ground,
by Definition \ref{def:must-summ} we know that \textit{(i)} $l\theta$ and
$l'\theta$ are also ground, and \textit{(ii)} $C_n|_{\BB} \models l\theta$
and $C_n|_{\BB} \models l'\theta$. This leads to a contradiction of our
assumption that $l\theta = \overline{l'}\theta$.
\end{myproof}
%%%%%%%%%%%%%%%%%%%%%%%%%%%%%%%%%%%%%%%%%%%%%%%%%%%%%%%%%%%%%%%%%%%%%%%%%%%
% FULL PROOF BELOW, ONLY SHOWN IF YOU COMPILE WITH SCRIPT ./makeWithProofs
%%%%%%%%%%%%%%%%%%%%%%%%%%%%%%%%%%%%%%%%%%%%%%%%%%%%%%%%%%%%%%%%%%%%%%%%%%%
\fi 
\begin{myproof}
We prove this by contradiction.
First, note that since $L^{mt}\theta$ is a set of ground literals
(by Definition \ref{def:must-summ} and because $e\theta$ is ground), if
$L^{mt}\theta$ is consistent, then for all literals $l,l' \in L^{mt}$
it is the case that $l\theta \neq \overline{l'}\theta$ (i.e., $l\theta$
is not the complement of $l'\theta$). Now let us assume that the theorem
does not hold. This means that there must exist a
%belief base $\BB_1$ and a
ground instance $e\theta$ of $e$, such that
%$\BB_1 \models \phi\theta$,
$\phi\theta\theta'$ is consistent for some ground substitution $\theta'$,
but $l\theta = \overline{l'}\theta$ for some $l,l' \in L^{mt}$.

%By Definition \ref{def:must-summ} both $l\theta$ and $l'\theta$ are ground.

Since $\phi\theta\theta'$ is consistent, there is at least one truth
assignment for its (ground) atoms that satisfies $\phi\theta\theta'$.
Suppose $\BB$ is the set of ground atoms consisting only of those
mentioned in $\phi\theta\theta'$ that were assigned the truth
value $\true$. Observe that $\BB \models \phi\theta\theta'$.
%
%of the ground atoms in $\phi\theta\theta'$
Then, according to Definition \ref{def:summ-prec} (Precondition),
%since $\BB \models \phi\theta$ holds,
there must exist a successful HTN execution $C_1 \cdot\ldots\cdot C_n$ of
$e\theta$ such that $C_1|_{\BB} = \BB$. Moreover, since $l,l'$ are must
literals of $e$ and $e\theta$ is ground, by Definition \ref{def:must-summ},
both $l\theta$ and $l'\theta$ are also ground. By the same definition,
$C_n|_{\BB} \models l\theta$ and $C_n|_{\BB} \models l'\theta$.
%However, our assumption states that $l\theta = \overline{l'}\theta$,
However, according to our assumption, $l\theta = \overline{l'}\theta$,
and therefore $C_n|_{\BB} \models \overline{l'}\theta$. This
contradicts the fact that $C_n|_{\BB} \models l'\theta$, because
$C_n|_{\BB}$ is ground due to our assumption in Section
``\nameref{sec:assum}'' that $\Pi$ is coherent.
\end{myproof}

In addition to must literals, there are two related 
notions. The first, called \textit{may summary conditions} 
in \cite{clement:jair:07}, defines literals that hold at the 
end of at least one successful \HTN\ execution of the program, 
and the second, weaker notion defines literals that are simply
mentioned in the program or in one of its ``descendant'' programs; 
such literals may or may not be brought about when the program executes.
It is this second notion, called \textit{mentioned literals}, 
that we use.

\begin{definition}[Mentioned Literal]\label{def:men-summ}
If $P$ is a program, its \defterm{mentioned literals} 
(relative to a plan-library $\Pi$ and an action-library 
$\Lambda$), denoted $mnt(P)$, is the set $mnt(P) = $ 
\[\small \left\{
\begin{array}{ll}
post(P,\Lambda)                                         & \mbox{if $P = +b \mid -b \mid act \mid\ ?\phi$},\\
mnt(P_1) \cup mnt(P_2)                                  & \mbox{if $P = P_1;P_2$},\\
\{l\theta'\ \mid\ \plane{e':\psi<-P'} \in \Pi,  
                                                        & \mbox{if $P =\ !e$}.\\
\ e = e'\theta,\ l \in mnt(P'),                           & \mbox{}\\
\ \theta'\ \text{is any substitution}\}                    & \mbox{}\\
\vspace{-10mm}
\end{array}
%\vspace{-6mm}
\right.\]\noindent
\end{definition}

We use this weaker notion because the stronger notion of 
a may summary condition in \cite{clement:jair:07} is not 
suitable for our approach, which reasons about plans that 
will not be interleaved with one another---i.e., plans that 
will be scheduled as a sequence.
%
%Let us illustrate this with an example. 
For example, consider the
figure below, which shows a plan-library for going 
to work on Fridays, possibly one
belonging to a larger library from an 
agent-based simulation. The expressions 
to the left and right sides of actions/plan-rules 
are their preconditions and postconditions, 
respectively.
% 
%\vspace{-2mm}
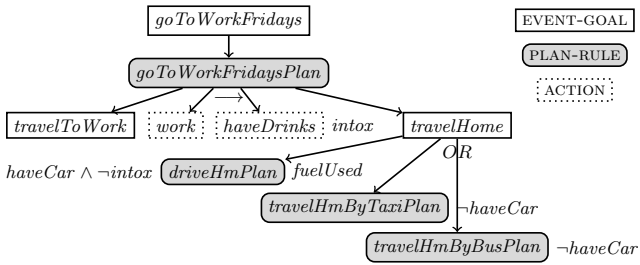
\begin{center}
\resizebox{1.03\linewidth}{!}{\begin{tabular}{c}
\begin{tikzpicture}[]
  \tikzstyle{every node}=[draw=black,thick]
  \tikzstyle{edge from parent}=[black,thick,draw,->]

  \tikzstyle{method}=[rounded corners, fill=black!15]
  \tikzstyle{photon} = [snake=snake, draw=red]
  \tikzstyle{cross} = [cross out,draw=blue,very thick,minimum size=0.8cm]
  \tikzstyle{ptask}=[dotted]

\tikzstyle{level 1}=[level distance=0.9cm]
\tikzstyle{level 2}=[level distance=0.9cm]
% child%[sibling distance=3cm] 

  \node (top) {$\GoToWorkFridays$}
      child {node[method,label=below:$\longrightarrow$] {$\GoToWorkFridaysPlan$}
		child[sibling distance=1.8cm]{node {$\TravelToWork$}}
                child[sibling distance=1.8cm]{node[ptask] {$\DoWork$}}
                child[sibling distance=1.4cm]{node[ptask,label=right:$\HadDrinks$] {$\HaveDrinks$}}
		child[sibling distance=2.6cm]{node[label=below:$OR$] {$\TravelHome$} 
			child[sibling distance=4.0cm, level distance=0.8cm] {node[method,label=left:$\HaveCar \land \neg \HadDrinks$,label=right:$\FuelUsed$] {$\DriveHomePlan$}}
                        child[sibling distance=1.5cm,level distance=1.42cm,xshift=-1.75cm] {node[method,,label=right:$\neg\HaveCar$]{$\TravelHomeByTaxiPlan$}}
                        child[sibling distance=0cm,level distance=2.1cm] {node[method,label=right:$\neg\HaveCar$]{$\TravelHomeByBusPlan$}}
		}	
	};

 \node [above right of=top,shift={(5.2cm,-0.70cm)}]{\sc event-goal};
 \node [method,above right of=top,shift={(5.2cm,-1.30cm)}]{\sc plan-rule};
 \node [ptask,above right of=top,shift={(5.2cm,-1.90cm)}]{\sc action};

\end{tikzpicture}
\end{tabular}

}
\end{center}
Observe that $\FuelUsed$ is actually never 
asserted in the context of the hierarchy shown, 
because literal $\neg \HadDrinks$ (``not
intoxicated'') in the context 
condition of $\DriveHomePlan$ is contradicted by 
literal $\HadDrinks$. However, the algorithms in 
\cite{clement:jair:07} will still classify 
$\FuelUsed$ as a may summary condition of 
plan $\GoToWorkFridaysPlan$, because some other 
plan may have a step asserting $\neg 
\HadDrinks$---perhaps a step that involves 
staying overnight in a hotel nearby---that 
can be ordered to occur between $\HaveDrinks$ 
and $\TravelHome$.

Since we cannot rely on such steps, we settle for 
a weaker notion---mentioned literals---than the 
corresponding definition of a may summary condition.
By our definition there can be literals that are 
mentioned in some plan-body but in fact can never 
be asserted, because of interactions that preclude
the particular plan-body which asserts that literal 
from being applied.
We avoid the approach of disallowing interactions
like the one shown above in order to use the 
stronger notion of a may summary condition 
because such interactions are natural in BDI and HTN domains: 
event-goals such as $\TravelHome$ are, intuitively, 
meant to be self-contained ``modules'' that can be 
``plugged'' into any relevant part of a hierarchical 
structure in order to derive all or just some of their 
capabilities. 

Finally, we conclude this section by combining the above 
definitions of must and mentioned literals to form the definition 
of the \textit{summary information} of a program.

\begin{definition}[Summary Information]\label{def:summ-infor}
If $P$ is a program, its \defterm{summary information} 
(relative to a plan-library and an action-library) is a tuple 
$\tuple{P,\phi,L^{mt},L^{mn}}$, where $\phi$ is a precondition
of $P$ if $P$ is an event-goal program, and $\phi = \epsilon$
otherwise; $L^{mt}$ is a set of must literals of $P$; and $L^{mn}$ 
is a set of mentioned literals of $P$.
\end{definition}

%%%%%%%%%%%%%%%%%%%%%%%%%%%%%%%%%%%%%%%%%%%%%%%%%%%%%%%%%%%%%%%%%%%%%%%%%%%%%%%%%%%%%
%%%%%%%%%%%%%%%%%%%%%%%%%%%%%%%%%%%%%%%%%%%%%%%%%%%%%%%%%%%%%%%%%%%%%%%%%%%%%%%%%%%%%
%%%%%%%%%%%%%%%%%%%%%%%%%%%%%%%%%%%%%%%%%%%%%%%%%%%%%%%%%%%%%%%%%%%%%%%%%%%%%%%%%%%%%

\section{EXTRACTING SUMMARY INFORMATION}
\label{sec:algos}

With the formal definitions now in place,
in this section we provide algorithms to 
extract summary information for event-goals
in a plan-library. Moreover, we illustrate 
the algorithms with an example, and analyse 
their properties.

Basically, we extract summary information
from a given plan-library and action-library 
by propagating up the summary information of 
lower-level programs, starting from the 
leaf-level ones in the plan-library, until 
we eventually obtain the summary information 
of all the top-level event-goals. 

\newcommand{\MustUndone}{\mbox{\textsf{Must-Undone}}}
\newcommand{\MayUndone}{\mbox{\textsf{May-Undone}}}

\begin{algorithm}[!t]
{\small
\textbf{Algorithm 1} \text{$\Summarise(\Pi, \Lambda)$}
%\vspace{0.5mm}
\hrule
\label{alg:summarise}
\begin{algorithmic}[1]

\REQUIRE Plan-library $\Pi$ and action-library $\Lambda$.

\ENSURE Set of summary info. of event-goal types in $\Pi$.

\STATE $\Delta \Leftarrow \{ \tuple{P, \epsilon, post(P,\Lambda), post(P,\Lambda)} \mid \newline
	P\ \text{is a primitive program mentioned in}\ \Pi \}$\label{alg:lin:prim-start}\\

\STATE $E \Leftarrow \{e(\vec{x}) \mid e\ \text{is an event-goal 
                       mentioned in}\ \Pi\}$ \label{alg:lin:s-e}  

\FOR[Recall $\RR_\Pi(e)$ is the rank of $e$]{$i \Leftarrow min(R)$ \textbf{to} $max(R)$ where \newline $R=\{ \RR_\Pi(e) \mid e \in E \}$} \label{alg:lin:lvl-start}

\FOR{each $e \in E$ such that $\RR_\Pi(e) = i$}\label{alg:lin:loop-start}

\STATE $\Delta \Leftarrow \Delta \cup \newline\{ \SummarisePlanBody(P, \Pi, \Lambda, \Delta) \mid 
	       \plane{e':\psi<-P} \in \Pi, e' = e\theta\}$\label{alg:lin:pr-start}\\

\STATE $\Delta \Leftarrow \Delta \cup \{ \SummariseEvent(e, \Pi, \Delta) \}$ \label{alg:lin:event-def} 
	
\ENDFOR
\ENDFOR \label{alg:lin:lvl-end}

\RETURN $\Delta \setminus \{u \mid u \in \Delta, \newline u\ \text{is not the summary information of an event-goal} \}$\label{alg:lin-last}

\end{algorithmic}
}
%\end{algorithm}[t]
%\begin{algorithm}[t]
{\small
\hrule
\hrule
\hrule
%\vspace{0.5mm}
\textbf{Algorithm 2} \text{$\SummarisePlanBody(P, \Pi, \Lambda, \Delta_{in})$}
%\vspace{0.5mm}
\hrule
\label{alg:planbodysumm}
\begin{algorithmic}[1]

\REQUIRE Plan-body $P$; plan-library $\Pi$;
         action-library $\Lambda$; and the set $\Delta_{in}$ of summary
         information of primitive programs and event-goal types mentioned in $P$.

\ENSURE The summary information of $P$.

\STATE $\Delta \Leftarrow \Delta_{in} \cup \{ \langle !e(\vec{x}), \phi, L^{mt},L^{mn} \rangle\theta \mid\ 
                                         !e(\vec{t})\ \text{occurs in}\ P, \newline
                                         \langle e(\vec{x}), \phi, L^{mt},L^{mn} \rangle \in \Delta_{in},
                                         e(\vec{t}) = e(\vec{x})\theta \}$\label{alg:lin:event-summ-start}

\COMMENT{We assume variables in $L^{mn}$ are appropriately renamed}

\STATE Let $P = P_1 ; P_2 ; \ldots ; P_n$ where each $P_i$ is atomic 

\STATE $L^{mt}_P \Leftarrow \{l \mid l \in L^{mt}, \tuple{P_i, \phi, L^{mt},L^{mn}} \in \Delta, \newline
                 i \in \{1,\ldots,n\}, \neg \MayUndone(l, P_{i+1};\ldots;P_n, \Delta)\}$\label{alg:lin:may-undone}

\STATE $L^{mn}_P \Leftarrow \{l \mid l \in L^{mt} \cup L^{mn}, \tuple{P_i, \phi, L^{mt},L^{mn}} \in \Delta, \newline
                i \in \{1,\ldots,n\}, \neg \MustUndone(l, P_{i+1};\ldots;P_n, \Delta)\}$\label{alg:lin:must-undone}

\RETURN $\tuple{P, \epsilon, L^{mt}_P, L^{mn}_P}$

\end{algorithmic}
}
%\end{algorithm}
%}
%\begin{algorithm}[t]
{\small
\hrule
\hrule
\hrule
%\vspace{0.5mm}
\textbf{Algorithm 3} \text{$\SummariseEvent(e(\vec{x}), \Pi, \Delta)$}
%\vspace{0.5mm}
\hrule
\label{alg:eventsumm}
\begin{algorithmic}[1]

\REQUIRE Event-goal type $e(\vec{x})$; plan-library $\Pi$;
         and the set $\Delta$ of summary
         information of plan-bodies of plan-rules $\plane{e':\psi<-P} \in
         \Pi$ such that $e' = e(\vec{x})\theta$.

\ENSURE The summary information of $e(\vec{x})$.

\STATE $\phi \Leftarrow \false, L^{mt} \Leftarrow \emptyset, L^{mn} \Leftarrow \emptyset,$ and $S \Leftarrow \emptyset$

%\STATE $L^{mt}, L^{mn}, S \Leftarrow \emptyset$

\COMMENT{$L^{mt},L^{mn}$ are sets of literals and $S$ is a set of sets of literals}

\FOR{each $\planev{e[y]:\psi<-P} \in \Pi$ such that $e(\vec{x}) =
     e(\vec{y})\theta$}\label{fig:lin:pre-sum-start}

\STATE $\phi \Leftarrow \phi \lor \psi\theta$ \label{alg:lin:disj}

\COMMENT{Relevant variables in $\psi$ and $L_P^{mt}, L_P^{mn}$ below are renamed}

\STATE $S \Leftarrow S \cup \{ L^{mt}_P\theta \}$, where $\tuple{P, \epsilon,
                               L^{mt}_P, L^{mn}_P} \in \Delta$\label{alg:lin:smust1}

\STATE $L^{mn} \Leftarrow L^{mn} \cup L^{mn}_P\theta$\label{alg:lin:smay}

\ENDFOR \label{fig:lin:pre-sum-end}

\IF[Obtain the must literals of $e(\vec{x})$]{$S \neq \emptyset$}

\STATE $L^{mt} \Leftarrow \bigcap S$\label{fig:lin:post-sum-end}

\STATE $L^{mt} \Leftarrow \{l \mid l \in L^{mt},\newline \text{variables occurring in}\ l\
               \text{also occur in}\ e(\vec{x}) \}$\label{fig:lin:post-sum-end2}

\ENDIF

\RETURN $\langle e(\vec{x}), \phi, L^{mt}, L^{mn} \rangle$

%\vspace{-1mm}
\end{algorithmic}
}
\end{algorithm}

To be able to identify must literals, 
we need to be able to determine 
whether a given literal is definitely undone, or 
\textit{must undone}, and possibly undone, or 
\textit{may undone} in a program.
Informally, a literal $l$ is must undone in a sequence 
$P$ of atomic programs if the literal's negation is a must 
literal of some atomic program in $P$. 
Formally, then, given a program $P$ and the set $\Delta$ of 
summary information of all atomic programs in $P$, 
a literal $l$ is \defterm{must undone} in $P$ relative to 
$\Delta$, denoted $\MustUndone(l,P,\Delta)$, if there exists
an atomic program $P'$ in $P$ and a literal 
$l' \in L^{mt}$, with 
$\tuple{P', \phi, L^{mt}, L^{mn}} \in \Delta$, 
such that $l = \overline{l'}$, that is, $l$ is the complement 
of $l'$.
%\footnote{The \textit{complement} of a literal 
%$l \in \{a, \neg a\}$ is $a$ if $l = \neg a$, and $\neg a$
%otherwise.}
 
Similarly, we can informally say that a literal $l$ is may 
undone in a program $P$ if there is a literal $l'$ that is a 
mentioned (or must) literal of some atomic program 
in $P$ such that $l'$ may become the negation of $l$ 
after variable substitutions.
\label{def:may-undone}
Formally, given a program $P$ and the set $\Delta$ of summary
information of all atomic programs in $P$, a literal $l$
is \defterm{may undone} in $P$ relative to $\Delta$, denoted
$\MayUndone(l,P,\Delta)$, if there exists an atomic program $P'$
in $P$, a substitution $\theta$, and a literal
$l' \in L^{mn}$,\footnote{variables occurring
in $l'$ are renamed to those not occurring in $l$} 
with $\tuple{P', \phi, L^{mt}, L^{mn}} \in \Delta$, such that 
$l\theta = \overline{l'}\theta$. 

\textbf{Algorithm 1.} %\ref{alg:summarise} 
This is the top-level algorithm for computing the summary information 
$\Delta$ of event-goal types occurring in the plan-library. 
The algorithm works bottom up, by summarising first the
leaf-level entities of the plan-library---the primitive programs
(line \ref{alg:lin:prim-start})---and then repetitively summarising 
plan-bodies (Algorithm 2) and event-goals
(Algorithm 3) in increasing order of their levels 
of abstraction (lines \ref{alg:lin:lvl-start}-\ref{alg:lin:lvl-end}).

\textbf{Algorithm 2.} This algorithm summarises the given plan-body $P$ by 
referring to the set $\Delta_{in}$ containing the 
summary information tuples of programs in $P$. First,
the algorithm obtains the summary information of each 
event-goal program in the plan-body from the summary 
information of the corresponding event-goal types in 
$\Delta_{in}$ (line \ref{alg:lin:event-summ-start}).
This involves substituting variables occurring
in relevant summary information tuples in 
$\Delta$ with the corresponding terms occurring 
in the event-goal program being considered.
Second, the algorithm computes the set of must literals ($L^{mt}_P$) and the
set of mentioned literals ($L^{mn}_P$) of the given plan-body $P$, by determining,
from the must and mentioned literals of atomic programs in $P$, which
literals will definitely hold and which ones will only possibly hold
on successful executions of $P$ (lines \ref{alg:lin:may-undone} and
\ref{alg:lin:must-undone}). More precisely, a must literal $l$ of an
atomic program $P_i$ in $P=P_1;\ldots;P_n$ is classified as a must
literal of $P$ only if $l$ is not may undone in $P_{i+1};\ldots;P_n$
(line \ref{alg:lin:may-undone}). Otherwise, $l$ is classified as 
only a mentioned literal of $P$, provided $l$ is not also must undone 
in $P_{i+1};\ldots;P_n$ (line \ref{alg:lin:must-undone}). The reason 
we do not summarise literals that are must undone is to avoid 
missing must literals in cases where they are possibly
undone but then later (definitely) reintroduced, as we
illustrate below.

Suppose, on the contrary, that the algorithm \textit{does}
summarise mentioned literals that are must undone. Then, 
given the plan-library below, the algorithm would 
(hypothetically) compute the summary information 
denoted by the two sets attached to each node, the 
one on the left being its set of must literals and 
the one on the right its set of mentioned literals.
 
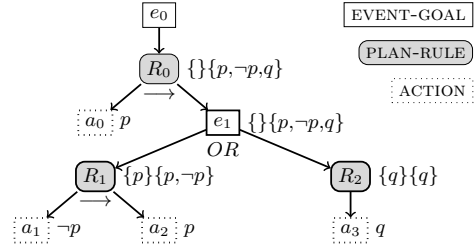
\begin{center}
\resizebox{0.75\linewidth}{!}{\begin{tabular}{c}
\begin{tikzpicture}[]
  \tikzstyle{every node}=[draw=black]
  \tikzstyle{edge from parent}=[black,thick,draw,->]

  \tikzstyle{method}=[rounded corners,fill=black!15]
  \tikzstyle{photon} = [snake=snake, draw=red]
  \tikzstyle{cross} = [cross out,draw=blue,very thick,minimum size=0.8cm]
  \tikzstyle{ptask}=[dotted]

\newcommand{\comma}{\mathit{,}}

%\tikzstyle{level 1}=[level distance=1cm, sibling distance=2cm]
%\tikzstyle{level 2}=[level distance=1cm, sibling distance=4cm]
% child%[sibling distance=3cm] 

  \node (top) {$e_0$}
      child[level distance=0.85cm] 
		{node[method,label=below:$\longrightarrow$,label=right:$\{\}\{p\comma \neg p\comma q\}$] {$R_0$} 	
		child[sibling distance=2cm]{node[ptask,label=right:$p$]{$a_0$}}
		child[sibling distance=2cm]{node[label=below:$OR$,label=right:$\{\}\{p\comma \neg p\comma q\}$] {$e_1$} 
			child[sibling distance=4.0cm] {node[method,label=below:$\longrightarrow$,label=right:$\{p\}\{p\comma \neg p\}$] {$R_1$}
				child[sibling distance=2.0cm] {node[ptask,label=right:$\neg p$] {$a_1$}}
				child[sibling distance=2.0cm] {node[ptask,label=right:$p$] {$a_2$}}
				}
                        child[sibling distance=4.0cm] {node[method,label=right:$\{q\}$$\{q\}$] {$R_2$}
                                child[sibling distance=1cm] {node[ptask,label=right:$q$] {$a_3$}}
                                %child[sibling distance=1cm] {node[ptask] {$a_4$}}
                                }
		}	
	};

\node [above right of=top,shift={(3.2cm,-0.70cm)}]{\sc event-goal};
\node [method,above right of=top,shift={(3.315cm,-1.30cm)}]{\sc plan-rule};
\node [ptask,above right of=top,shift={(3.575cm,-1.90cm)}]{\sc action};

\end{tikzpicture}
\end{tabular}

}
\end{center}
 
Observe that literal $p$ asserted by $a_0$ is 
not recognised as a must literal of $R_0$ simply
because it is may undone by mentioned literal $\neg p$ 
of $e_1$ (asserted by $a_1$), despite the fact 
that action $a_2$ of $R_1$ also subsequently adds $p$. 
On the other hand, our algorithm does recognise $p$ 
as a must literal of $R_0$ by not including $\neg p$ 
in the set of mentioned literals of $R_1$ (line 
\ref{alg:lin:must-undone}).
%because it is must undone in $R_1$. 

\textbf{Algorithm 3.} This algorithm summarises the 
given event-goal type $e(\vec{x})$ by referring to the set $\Delta$ containing 
the summary information tuples associated with the plan-bodies of 
plan-rules handling $e(\vec{x})$. In 
lines \ref{fig:lin:pre-sum-start} and 3, the algorithm 
takes the precondition of the event-goal 
as the disjunction of the context conditions of all associated 
plan-rules.\footnote{We do not need to ``propagate up'' context 
conditions as we do with plan-bodies' summary information because 
higher-level context conditions account for lower-level 
ones due to Definition \ref{def:safeLib}.} 
Then, the algorithm obtains the must and mentioned literals of the event-goal 
by respectively taking the intersection of the must literals of associated 
plan-rules (lines \ref{alg:lin:smust1} and \ref{fig:lin:post-sum-end}),
and the union of the mentioned literals of associated plan-rules 
(line \ref{alg:lin:smay}).
Applying substitution $\theta$ 
in line \ref{alg:lin:smust1} helps recognise
must literals of $e(\vec{x})$, by ensuring
that variables occurring in the summary 
information of its associated plan-bodies 
have consistent names with respect to $e(\vec{x})$.

%%%%%%%%%%%%%%%%%%%%%%%%%%%%%%%%%%%%%%%%%%%%%%%%%%%%%%%%%%%%%%%%%%%%%%%%%%%%
%%%%%%%%%%%%%%%%%%%%%%%%%%%%%%%%%%%%%%%%%%%%%%%%%%%%%%%%%%%%%%%%%%%%%%%%%%%%
%%%%%%%%%%%%%%%%%%%%%%%%%%%%%%%%%%%%%%%%%%%%%%%%%%%%%%%%%%%%%%%%%%%%%%%%%%%%

\subsection{An illustrative example}
\label{sec:example}

We shall illustrate the three algorithms with 
the example of a simple agent (like the ones
in \cite{ceballos2011genom}) exploring the
surface of Mars. A part of the agent's 
domain is depicted as a hierarchy in Figure 
\ref{fig:summ-info-table2}. The hierarchy's
top-level event-goal is to explore a given 
soil location $Y$ from current location $X$. 
This is achieved by plan-rule 
$R_0$, which involves navigating to the 
location and then doing a soil experiment. 
Navigation is achieved by rules $R_1$ and
$R_2$, which 
involve moving to the location, possibly 
after calibrating some of the rover's 
instruments. Doing a soil experiment 
involves the two sequential event-goals
of getting soil results for $Y$ and 
transmitting them to the lander. 
Specifically, the former is refined into 
actions such as determining moisture 
content and average soil particle size, 
and transmitting results involves either
establishing a connection with the lander, 
sending it the results, and then terminating 
the connection, or if the lander is not within 
range, navigating to it and uploading the soil results.
The table in Figure \ref{fig:summ-info-table2} shows 
the summary information computed by our 
algorithms for elements in the
figure's hierarchy.
%Figure \ref{fig:summ-info-table2}. 
Below, we describe 
some of the more interesting values in
the table.

\begin{figure*}[t]
\hspace{-2mm}
\resizebox{0.58\linewidth}{!}{\begin{tabular}{c}
\begin{tikzpicture}[]
  \tikzstyle{every node}=[draw=black]
  \tikzstyle{edge from parent}=[black,thick,draw,->]

  \tikzstyle{ptask}=[dotted]
  \tikzstyle{method}=[rounded corners, fill=black!15]
  \tikzstyle{photon} = [snake=snake, draw=red]
  \tikzstyle{cross} = [cross out,draw=blue,very thick,minimum size=0.8cm]
  \tikzstyle{ptask}=[dotted]

  %\tikzstyle{method}=[rounded corners]
  %\tikzstyle{photon} = [snake=snake, draw=red]
  %\tikzstyle{cross} = [cross out,draw=blue,very thick]%,minimum size=0.8cm]

 \tikzstyle{level 0}=[level distance=0.9cm, sibling distance=5cm]
 \tikzstyle{level 1}=[level distance=0.9cm, sibling distance=5cm]
 \tikzstyle{level 2}=[level distance=0.9cm, sibling distance=5cm]
 \tikzstyle{level 3}=[level distance=0.9cm, sibling distance=1.5cm]
 \tikzstyle{level 4}=[level distance=0.9cm, sibling distance=1.5cm]
 \tikzstyle{level 5}=[level distance=0.9cm, sibling distance=1.5cm]
 \tikzstyle{level 6}=[level distance=0.9cm, sibling distance=1.5cm]

  %\fontsize{9pt}{9pt}\selectfont
  \node (top) {$\ExploreSoilLocation(X,Y)$}
      child{node[method,label=below:$\longrightarrow$] {$R_0$} 	
		child[sibling distance=5.0cm] {node[label=below:$or$] {$\Navigate(X,Y)$} 
			child[sibling distance=2cm] {node[method,label=below:$\longrightarrow$] {$R_1$}
				child[sibling distance=1.7cm] {node[ptask] {$\CalibrateViaGPS$}}
				child[sibling distance=1.6cm] {node[ptask] {$\Move(X,Y)$}}
				}
                        child[sibling distance=3.9cm] {node[method] {$R_2$}
                                child {node[ptask] {$\Move(X,Y)$}}
                                }
		}	
		child[sibling distance=2.8cm]
			{node {$\PerformSoilExperiment(Y)$}
			child[level distance=2.0cm] {node[method,label=below:$\longrightarrow$] {$R_3$}
				child[sibling distance=8.8cm] {node {$\ObtainSoilResults(Y)$}
                                        child {node[method,label=below:$\longrightarrow$] {$R_4$}            
                                                child[sibling distance=2.35cm]{node[ptask] {$\PickSoilSample(Y)$}}
                                                child[sibling distance=2.35cm]{node {$\AnalyseSoilSample(Y)$}
                                        		child{node[method,label=below:$\longrightarrow$] {$R_5$}            
                                                		child[level distance=1cm,sibling distance=2.6cm] {node[ptask] {$\GetMoistureContent(Y)$}}
                                                		child[level distance=1cm,sibling distance=2.6cm] {node[ptask] {$\GetSoilParticleSize(Y)$}}
                                                		%child[sibling distance=4cm] {node[ptask] {$\GetSoilDensity(Y)$}}
                                        		} 
						} 
                                                child[sibling distance=2.35cm] {node[ptask] {$\DropSoilSample(Y)$}}
                                        }  
				}
				child[sibling distance=4.2cm] {node[label={[anchor=west,yshift=-1.5mm,xshift=-4.7mm]below:$or$}] {$\TransmitSoilResults(Y)$}
					child[level distance=1.8cm, sibling distance=4.7cm] {node[method,label=below:$\longrightarrow$] {$R_6$}
						child[sibling distance=2.15cm] {node[ptask] {$\EstablishConnection$}}
						child[sibling distance=1.9cm] {node[ptask] {$\SendResults(Y)$}}
						child[sibling distance=1.9cm] {node[ptask] {$\BreakConnection$}}
					}
                                        child[sibling distance=0.1cm] {node[method,label=below:$\longrightarrow$] {$R_7$}
                                                child[sibling distance=1.6cm] {node {$\Navigate(Y,L)$}}
                                                child[sibling distance=2.4cm] {node[ptask] {$\UploadResults(Y)$}}
                                        }  
				}
			}
		}
	};
 \node [above right of=top,shift={(-5.3cm,-0.7cm)}]{\sc event-goal};
 \node [method,above right of=top,shift={(-5.3cm,-1.3cm)}]{\sc plan-rule};
 \node [ptask,above right of=top,shift={(-5.3cm,-1.8cm)}]{\sc action};
 %\node [above right of=t4,shift={(2.5cm,-1cm)}]
 %          [text centered,rounded corners]
 %       { \begin{tabular}{ccc}
 %		Action & Prec. & Post. \\ \hline
 %		$a_1$  & $p$ & $q$ \\
 %		$a_2$  & $q$ & $r$ \\
 %		$a_3$  & $r$ & $s$ \\
 %		$a_4$  & $p$ & $q$ \\
 %		$a_5$  & $q$ & $r$ \\
 %		$a_6$  & $s$ & $t$ \\
 %	  \end{tabular}
 % 	};

\end{tikzpicture}
\end{tabular}}
%\caption{Part of a simple Mars Rover domain}
%\label{fig:summ-info}
%\end{figure}
%\qquad
%\begin{table}
%\resizebox{1.025\linewidth}{!}{\input{figures/mars-rover.tex}}
\hspace{-3mm}
{\footnotesize
\begin{tabular}[p]{|l|l|l|} \hline
\lft Program                   & Must Literals                                   & Mentioned Literals\\\hline
\lft$\CalibrateViaGPS$              & $\CCA$                                               & -     \\
\lft$\Move(X, Y)$             & $\neg\LAt(X), \LAt(Y)$                          & -     \\
\lft$\PickSoilSample(Y)$         & $\HSS(Y)$                                          & -     \\
\lft$\DropSoilSample(Y)$         & $\neg\HSS(Y)$                                      & -     \\
\lft$\GetMoistureContent(Y)$  \hspace{-8mm}   & $\HMC(Y)$                                          & -     \\
\lft$\GetSoilParticleSize(Y)$    & $\HPS(Y)$                                          & -     \\
\lft$\EstablishConnection$          & $\CE$                                                 & -     \\
\lft$\SendResults(Y)$            & $\RT(Y)$                                           & -     \\
\lft$\BreakConnection$              & $\neg\CE$                                             & -     \\
\lft$\UploadResults(Y)$          & $\RT(Y)$                                           & -     \\ \hline
\lft$P_1$                          & $\neg\LAt(X), \LAt(Y), \CCA$ \hspace{-8mm}                     & -     \\
\lft$P_2$                          & $\neg\LAt(X),\LAt(Y)$                         & -     \\
\lft$P_5$                          & $\HMC(Y), \HPS(Y)$                             & -     \\
\lft$P_4$                          & $\HMC(Y), \HPS(Y),$ \hspace{-8mm}                      & -     \\
                               & $\neg\HSS(Y)$                                  & -     \\
\lft$P_6$                          & $\RT(Y), \neg\CE$                              & -     \\
\lft$P_7$                          & $\neg\LAt(Y), \LAt(L),$                        & $\CCA$     \\
                               & $\RT(Y)$                                       &            \\
\lft$P_3$                          & $\RT(Y), \HMC(Y),$                             & $\neg\CE, \neg\LAt(Y), \LAt(L),\hspace{-2.0mm}$ \\
                                & $\HPS(Y), \neg\HSS(Y)$                                      & \CCA\\
\lft$P_0$                          & $\RT(Y), \HMC(Y),$              & $\neg\CE, \LAt(Y), \neg\LAt(Y),\hspace{-2.0mm}$ \\
                                & $\HPS(Y), \neg\HSS(Y)$                                       & $\LAt(L), \CCA, \neg\LAt(X)$ \\
 \hline
\lft$\Navigate(X, Y)$        & $\neg\LAt(X),\LAt(Y)$                           & $\CCA$ \\
\lft$\AnalyseSoilSample(Y)$     & Same as $P_5$                                         & - \\
\lft$\ObtainSoilResults(Y)$     & Same as $P_4$                                         & - \\
\lft$\TransmitSoilResults(Y)$ \hspace{-8mm}  & $\RT(Y)$                                            & $\neg\CE, \neg\LAt(Y), \LAt(L),\hspace{-2.0mm}$\\ 

&    & \CCA\\ 

\lft$\PerformSoilExperiment(Y)$ & Same as $P_3$                                         & Same as $P_3$ \\
\lft$\ExploreSoilLocation(X, Y)$   & Same as $P_0$                                         & Same as $P_0$ \\ \hline
\end{tabular}
%\vspace{1mm}
\caption{Must and mentioned literals (right) of atomic programs and 
	 plan-bodies in the hierarchy (left). The 
	 rightmost column only shows mentioned literals 
	 that are not also must literals. Abbreviations 
	 in the table are as follows: 
	 $\CCA = \Calibrated, \HSS = \HaveSoilSample, \HMC = 
	  \HaveMoistureContent, \HPS = \HaveParticleSize, \CE 
	  = \ConnectionEstablished, \RT = \ResultsTransmitted,$
	 and variable $L = Lander$. Rule $R_7$'s context 
	 condition binds $L$ to the lander's location. Each 
	 plan-body $P_i$ corresponds to rule $R_i$ in the 
	 hierarchy.}
\label{fig:summ-info-table2}
}
%\end{table}
\end{figure*}

\textbf{Plan-body $P_7.$} Must literals 
$\neg\LAt(Y)$ and $\LAt(L)$ of $P_7$ are 
derived from those of $\Navigate(X,Y)$,
after renaming variables $X$ and $Y$ to 
respectively $Y$ and $L$ in line 1 of 
Algorithm 2.

\textbf{Plan-body $P_4.$}
While $\HSS(Y)$ is a must literal 
of $P_4$'s primitive action 
$\PickSoilSample(Y)$, the literal 
is must undone by $P_4$'s last 
primitive action $\DropSoilSample(Y)$. 
Thus, $\HSS(Y)$ is not a must
(nor mentioned) literal of $P_4$. On 
the other hand, literal $\neg\HSS(Y)$ 
is indeed a must literal of $P_4$, along 
with literals $\HMC(Y)$ and $\HPS(Y)$, 
both of which are derived from the 
summary information of event-goal 
$\AnalyseSoilSample(Y)$.
 
\textbf{Plan-body $P_0.$}
%
%Observe that literal $\Calibrated$ is only 
%a mentioned literal of $P_0$ because the 
%literal is a mentioned literal of event-goal 
%$\Navigate(X,Y)$ and it is not must undone 
%in $\PerformSoilExperiment(Y)$.
%
While $\neg\LAt(X)$ is a 
must literal of event-goal $\Navigate(X,Y)$, 
it is only a mentioned literal of 
$P_0$ because it is may undone in event-goal
$\PerformSoilExperiment(Y)$; specifically, 
its mentioned literal $\LAt(L)$ is such that
$\neg\LAt(X)\theta = \neg\LAt(L)\theta$ for
$\theta=\{X/L\}$. Similarly, must 
literal $\LAt(Y)$ of $\Navigate(X,Y)$
is also may undone in $\PerformSoilExperiment(Y)$.

\textbf{Event-goal} $\TransmitSoilResults(Y).$
Since literal $\RT(Y)$ is a must 
literal of both of the event-goal's 
associated plan-bodies $P_6$ and 
$P_7$, and $Y$ also occurs in the 
event-goal, the literal is classified 
as a must literal of the event-goal. 
Recall that this means that for any 
ground instance 
$\TransmitSoilResults(Y)\theta$ of the 
event-goal, literal $\RT(Y)\theta$ 
holds at the end of any 
successful \HTN\ execution of 
$\TransmitSoilResults(Y)\theta$.

%irrespective of which of $R_6$ 
%or $R_7$ is chosen as the 
%refinement.

%%%%%%%%%%%%%%%%%%%%%%%%%%%%%%%%%%%%%%%%%%%%%%%%%%%%%%%%%%%%%%%%%%%%%%%%%%%%
%%%%%%%%%%%%%%%%%%%%%%%%%%%%%%%%%%%%%%%%%%%%%%%%%%%%%%%%%%%%%%%%%%%%%%%%%%%%
%%%%%%%%%%%%%%%%%%%%%%%%%%%%%%%%%%%%%%%%%%%%%%%%%%%%%%%%%%%%%%%%%%%%%%%%%%%%

\subsection{Soundness and Completeness}
\label{sec:correct}

We shall now analyse the properties of the 
algorithms presented. We show 
that they are sound, and we then discuss 
completeness. 
First, it is not difficult to see that the presented 
algorithms terminate, and that they run in polynomial time.

\begin{theorem}
\label{thm:summ-alg-complexity}
Algorithm 1 always terminates, and runs in polynomial 
time on the number of symbols occurring in $\Pi \cup \Lambda$.
\end{theorem}
\ifdefined\withsketch
\begin{myproof}
Since the algorithms presented are non-recursive, the only non-trivial
part of the proof concerns the procedure for computing a unification
when determining whether $\MayUndone(l,P,\Delta)$ holds (for a literal
$l$, program $P$, and a set $\Delta$ of summary information). In
\cite{Martelli:1982}, one such unification procedure is presented that
is linear on the number of symbols occurring in the two literals to
be unified.
\end{myproof}
\fi
%
%%%%%%%%%%%%%%%%%%%%%%%%%%%%%%%%%%%%%%%%%%%%%%%%%%%%%%%%%%%%%%%%%%%%%%%%%%%
% FULL PROOF BELOW, ONLY SHOWN IF YOU COMPILE WITH SCRIPT ./makeWithProofs
%%%%%%%%%%%%%%%%%%%%%%%%%%%%%%%%%%%%%%%%%%%%%%%%%%%%%%%%%%%%%%%%%%%%%%%%%%%
%
\begin{myproof}
Let $n_e$ be the total number of event-goal types occurring in $\Pi$
and $n=1$. Since a ranking function does exist for $\Pi$, we can
rank it as follows. For each event-goal type $e$ occurring in $\Pi$
that does not also occur in a plan-body mentioned in $\Pi$, we first
set $n$ to $n + n_e$, and then recursively assign the rank $n$ to
$e$ and $n-1$ to its children event-goal types that are either not
already ranked or have a higher rank.

Since the remaining algorithms are not recursive, the only non-trivial
part is the algorithm for computing a unification in $\MayUndone(l,P,\Delta)$
(where $l$ is a literal, $P$ is a program and $\Delta$ is a set of summary
information). This was shown to be linear on the number of symbols
occurring in the two literals to be unified \cite{Martelli:1982}.
\end{myproof}

This result is important when the plan-library 
changes over time, e.g. because the agent learns from past
experience, and summary information needs to be recomputed
frequently, or when it needs to be computed right at the
start of HTN planning, as done in \cite{tsuneto:98}.

The next result states that whenever Algorithm 1 %\ref{alg:summarise} 
($\Summarise$) classifies a literal as a must literal of an event-goal, 
this is guaranteed to be the case, and that the algorithm correctly 
computes its precondition and mentioned literals. More specifically,
any computed tuple, which includes one event-goal type $e$, formula 
$\phi$ and must literals $L^{mt}$, respects Definitions \ref{def:summ-prec} 
and \ref{def:must-summ}. Moreover, there is exactly one tuple associated
with $e$.

\begin{theorem}\label{thm:summ-alg-correctness2}
Let $\Pi$ be a plan-library, $\Lambda$ be an action-library, $e$ be an event-goal 
type mentioned in $\Pi$, and let $\Delta_{out} = \Summarise(\Pi,\Lambda)$. 
There exists one tuple $\tuple{e, \phi, L^{mt}, L^{mn}} \in \Delta_{out}$,
%such that the tuple is the summary information of $e$, and $L^{mn} \subseteq mnt(e)$
the tuple is the summary information of $e$, and $L^{mn} \subseteq mnt(e)$
(recall $mnt(e)$ denotes the mentioned literals of $e$).
%is a set of RML of $e$. 
%
\end{theorem}
\ifdefined\withsketch
\begin{myproof}
We prove this by induction on $e$'s rank in $\Pi$. First,
from our ranking function we obtain a new one $\RR_\Pi$ by
making event-goal ranks ``contiguous'' and start from $0$.
  
For the \textit{base case}, take any event-goal $e$
with $\RR_\Pi(e) = 0$.
According to Definition \ref{def:ranking} (Ranking),
if $\RR_\Pi(e) = 0$, then $children(e,\Pi) = 
\emptyset$. Thus, for all rules $\plane{e':\psi<-P} 
\in \Pi$ such that $e = e'\theta$, no
event-goals occur in $P$. Let $P_{all}$ 
be the set of plan-bodies of all such rules.
Then, the two main steps are as follows.
First, we show that due to line \ref{alg:lin:prim-start}
of procedure $\Summarise(\Pi,\Lambda)$ there is exactly
one summary tuple
$\tuple{P', \epsilon, L_{P'}^{mt}, L_{P'}^{mn}} \in \Delta$
for each primitive program $P'$ mentioned in each $P \in P_{all}$.
Second, since 
$\SummarisePlanBody(P,\Pi,\Lambda,\Delta)$ is called in line
\ref{alg:lin:pr-start} for each plan-body $P \in P_{all}$, we
show that on the completion of this line, there is exactly
one summary tuple
$\tuple{P, \epsilon, L_P^{mt}, L_P^{mn}} \in \Delta$ for
each plan-body $P \in P_{all}$. A similar argument applies
to line \ref{alg:lin:event-def}.
  
For the \textit{induction hypothesis}, we assume that the 
theorem holds if $\RR_\Pi(e) \leq k$, for some $k \in 
\mathbb{N}_0$.
  
For the \textit{inductive step}, we show that the theorem
holds for $\RR_\Pi(e) = k+1$. The main steps are as follows.
Let $E_{all}$ denote the (non-empty) set of event-goal 
types mentioned in all plan-bodies $P \in P_{all}$,
where $P_{all}$ is as before.
By Definition \ref{def:ranking}, we know that for all
$e' \in E_{all}$, $\RR_\Pi(e') < \RR_\Pi(e)$. Then,
by the induction hypothesis, it follows that for 
each $e' \in E_{all}$, there is exactly one tuple 
$\tuple{e', \phi_{e'}, L_{e'}^{mt}, L_{e'}^{mn}} 
\in \Delta$, and this tuple is the summary
information of $e'$.
Finally, since $e$ has a higher rank than those 
of all event-goals $e' \in E_{all}$, 
$\Summarise(\Pi,\Lambda)$ will only call 
$\SummariseEvent(e,\ldots)$ after the above 
tuples are added to $\Delta$, resulting in 
tuple $\tuple{e, \phi, L^{mt}, L^{mn}}$ 
also being added to $\Delta$.
\end{myproof}
\fi
%
%%%%%%%%%%%%%%%%%%%%%%%%%%%%%%%%%%%%%%%%%%%%%%%%%%%%%%%%%%%%%%%%%%%%%%%%%%%
% FULL PROOF BELOW, ONLY SHOWN IF YOU COMPILE WITH SCRIPT ./makeWithProofs
%%%%%%%%%%%%%%%%%%%%%%%%%%%%%%%%%%%%%%%%%%%%%%%%%%%%%%%%%%%%%%%%%%%%%%%%%%%
%
\begin{myproof}
%
% I'VE COMMENTED OUT THE FOLLOWING AS IT SEEMED LIKE OVERKILL - Lavindra
% There's a simpler paragraph below.
%We prove this by induction on the rank of $e$ in $\Pi$. First, from the
%ranking function $\RR_\Pi$ we obtain the ``contiguous'' ranking function
%%
%\[
%\begin{array}{lll}
%\RR'_\Pi &      =       & \{(e^0_1,0),\ldots,(e^0_i,0),\\
%         &              & \phantom{\{} (e^1_1,1),\ldots,(e^1_j,1),\\
%         &              & \phantom{\{} \ldots,\\
%         &              & \phantom{\{} (e^n_1,n),\ldots,(e^n_k,n)\}.
%\end{array}
%\]
%such that if $\RR_\Pi \neq \emptyset$, then \textit{(i)} $i \geq 1$ and
%\textit{(ii)} for any event-goals $e,e'$, there exist $(e,a),(e',a) \in \RR'_\Pi$
%for some $a \in \mathbb{N}_0$ if and only if there exist $(e,b),(e',b) \in \RR_\Pi$
%for some $b \in \mathbb{N}_0$. Observe that it is straighforward to obtain
%such a ranking function $\RR'_\Pi$. Henceforth we refer to $\RR'_\Pi$ as
%$\RR_\Pi$.\\
%
We prove this by induction on $e$'s rank in $\Pi$. First,
from our ranking function we obtain a new one $\RR_\Pi$ by
making event-goal ranks ``contiguous'' and start from $0$.\\

\noindent{\bf [Base Case]} Let $e$ be an event of rank $0$ in $\Pi$,
that is, $\RR_\Pi(e) = 0$.
Observe from the definition of a ranking for a plan-library
(Definition \ref{def:ranking}) that if $\RR_\Pi(e) = 0$, then
$children(e,\Pi) = \emptyset$. This entails that for all
plan-rules $\plane{e':\psi<-P} \in \Pi$ such that $e = e'\theta$,
no event-goals are mentioned in $P$. There are two cases to consider.

\textit{Case 1.1.}
In the special case where no such plan-rule exists, then the call to
procedure $\SummariseEvent(e,\Pi,\Delta)$ in line \ref{alg:lin:event-def}
of procedure $\Summarise(\Pi,\Lambda)$ returns tuple $\tuple{e, \false,
\emptyset, \emptyset}$, which is indeed the summary information of $!e$.

\textit{Case 1.2.}
Now consider the case where there \textit{are} one
or more plan-rules in $\Pi$ such that for all $\plane{e':\psi<-P} \in \Pi$
it is the case that $e = e'\theta$ but no event-goals are mentioned in $P$.
Let $P_{all}$ denote the (non-empty) set of all such plan-bodies.
Then, we know from Lemma \ref{thm:atomic-prog-summ-correctness} that,
due to line \ref{alg:lin:prim-start} in the algorithm, there is
exactly one tuple $\tuple{P', \epsilon, L_{P'}^{mt}, L_{P'}^{mn}} \in
\Delta$ for each primitive program $P'$ mentioned in each plan-body
$P \in P_{all}$, such that the tuple is the summary information of
$P'$.

Next, observe that before reaching
line \ref{alg:lin:event-def} of procedure $\SummariseEvent(\Pi,\Lambda)$,
procedure $\SummarisePlanBody(P,\Pi,\Lambda,\Delta)$ is called
in line \ref{alg:lin:pr-start} for each plan-body $P \in P_{all}$.
Then, from Lemma \ref{thm:planbodysumm-alg-correctness2}, we know that,
on the completion of line \ref{alg:lin:pr-start}, there is exactly one tuple
$\tuple{P, \epsilon, L_P^{mt}, L_P^{mn}} \in \Delta$ for each plan-body
$P \in P_{all}$ such that the tuple is the summary information of $P$.
Finally, from Lemmas \ref{thm:eventsumm-alg-correctness2} and
\ref{thm:eventsumm-alg-correctness3}, we can conclude that on the
completion of line \ref{alg:lin:event-def} of the algorithm
(i.e., after calling procedure
$\SummariseEvent(e,\Pi,\Delta)$), there is exactly
one tuple $\tuple{e, \phi_{e}, L_e^{mt}, L_e^{mn}} \in \Delta$ such
that the tuple is the summary information of $e$.
Therefore, the theorem holds for the base case.\\

\noindent{\bf [Induction Hypothesis]} Assume that the theorem holds
if $\RR_\Pi(e) \leq k$, for some $k \in \mathbb{N}_0$.\\

\noindent{\bf [Inductive Step]} Suppose $\RR_\Pi(e) = k+1$. Let
$P_{all} = \{P \mid \plane{e':\psi<-P} \in \Pi, e = e\theta\}$.
There are two cases to consider.

\textit{Case 2.1.}
First, there is no plan-body
$P \in P_{all}$ such that there is an event-goal mentioned in $P$
(i.e., all plan-bodies in $P_{all}$ mention only primitive programs).
Thus, $children(e,\Pi) = \emptyset$. %(Definition \ref{def:children}).
If $P_{all} \neq \emptyset$, then the proof for this case is the
same as \textit{Case 1.2} in the Base Case above.

\textit{Case 2.2.}
On the other hand, if $P_{all} = \emptyset$, then
the proof for this case is the same as \textit{Case 1.1} in the Base Case above.

\textit{Case 2.3.}
The third case is that $P_{all} \neq \emptyset$ and there exists a
plan-body $P \in P_{all}$ such that an event-goal is
mentioned in $P$.
Then, let $E_{all}$ denote the (non-empty) set of event-goal types of
all event-goals mentioned in all plan-bodies $P \in P_{all}$.
From Definition \ref{def:ranking} (Ranking), for all event-goals $e' \in E_{all}$,
$\RR_\Pi(e') < \RR_\Pi(e) \leq k$. Then, by the induction hypothesis, for each
$e' \in E_{all}$, there is exactly one tuple $\tuple{e', \phi_{e'}, L_{e'}^{mt},
L_{e'}^{mn}} \in \Delta_{out}$ such that the tuple is the summary information of $e'$.
It is not difficult to see from procedure $\Summarise$
that all such tuples exist in $\Delta_{out}$ because the value returned
by procedure $\SummariseEvent(e',\Pi,\Delta)$ is added to set $\Delta$
in line \ref{alg:lin:event-def}, for each event-goal $e' \in E_{all}$.
After this, since $e$ has a higher rank than all $e' \in E_{all}$,
procedure
$\Summarise(\Pi,\Lambda)$ will repeat its outer loop at least one
more time (note that there may be other event-goals mentioned in $\Pi$
with the same rank as $e$ or higher).
The procedure will then call $\SummariseEvent(e,\ldots)$, by which
point procedure $\SummarisePlanBody(P,\ldots)$ will have already
been called for each plan-body $P \in P_{all}$, and in turn, each
such call will only have occurred after procedure
$\SummariseEvent(e',\ldots)$ is called for each
event-goal $e' \in E_{all}$.

Then, by Lemmas \ref{thm:atomic-prog-summ-correctness} and by
\ref{thm:planbodysumm-alg-correctness2} and the induction hypothesis,
it follows that on the completion of the call to $\SummarisePlanBody(P,\ldots)$
in line \ref{alg:lin:pr-start}
for each $P \in P_{all}$, there is exactly one tuple $\tuple{P, \epsilon,
L_P^{mt}, L_P^{mn}} \in \Delta$ such that the tuple is the summary
information of $P$.
Finally, from Lemmas \ref{thm:eventsumm-alg-correctness2}
and \ref{thm:eventsumm-alg-correctness3}, we can conclude that after
calling procedure $\SummariseEvent(e,\Pi,\Delta)$ in line \ref{alg:lin:event-def},
there is exactly one tuple $\tuple{e, \phi_{e}, L_e^{mt}, L_e^{mn}} \in \Delta$ such
that the tuple is the summary information of $e$.
Therefore, the theorem holds.
\end{myproof}

Next, we discuss completeness. The theorem
below states that any precondition computed
by Algorithm 3 is complete: i.e., given any 
state from where there is a successful \HTN\ 
execution of an event-goal, the precondition
extracted for the event-goal will hold in 
that state. This theorem only concerns 
Algorithm 3 because we can compute 
preconditions of event-goals without 
needing to compute preconditions of plans.

\begin{theorem}\label{thm:summ-alg-correctness2}
Let $\Pi$ be a plan-library, $\Lambda$ an action-library, $e$
an event-goal type mentioned in $\Pi$, and let
$\tuple{e, \phi, L^{mt}, L^{mn}} \in \Summarise(\Pi,\Lambda)$.
For all ground instances $!e\theta$ of $!e$ and belief bases
$\BB$ such that there exists a successful \HTN\ execution
$C_1 \cdot\ldots\cdot C_n$ of $!e\theta$ with $C_1|_{\BB} = \BB$,
it is the case that $\BB \models \phi\theta$.
\end{theorem}
\begin{myproof}
If $!e\theta$ has a successful \HTN\ 
execution, then there is
also a plan-rule $\plane{e':\psi<-P} 
\in \Pi$ associated with $e$ such that
$\BB \models \psi'\theta$ holds, where
$\psi'$ is an appropriate renaming of
variables in $\psi$. Since $\psi'$ is a 
disjunct of $\phi$ (line 3 of Algorithm 3), 
it follows that $\BB \models \phi\theta$.
(See also proof of Lemma 9.)
\end{myproof}

There are, however, situations where 
the algorithms do not detect all 
\textit{must literals} of an event-goal. 
The underlying reason for this is that 
we do not reason about (FOL) precondition 
formulas; specifically, we do not check 
entailment, because this is semi-decidable 
in general \cite{gabbay:book}. In what
follows, we use examples to characterise 
the four cases in which the algorithms
are unable to recognise must literals, 
and show how some of the cases can be 
averted.

The first case was depicted
in our example about 
going to work on Fridays: by Definition 
\ref{def:must-summ}, literal $\neg\HaveCar$ 
is a must literal of $\GoToWorkFridaysPlan$, 
but Algorithm 2 classifies it as only a 
mentioned literal, as
it cannot infer that the context condition 
of rule $\DriveHomePlan$ is contradicted by 
literal $\HadDrinks$, and therefore that 
$\DriveHomePlan$ can never be applied. 

The second case is where a literal
is a must literal simply because
it is entailed by a context condition.
For example, take an event-goal 
$mov(P,T,L)$ that is associated with
one plan-rule, whose context condition
checks whether package $P$ is in truck 
$T$, i.e., $\IN(P,T)$, and whose plan-body 
moves the truck to location $L$. Observe that
$\IN(P,T)$ is a must literal of $mov(P,T,L)$ 
by definition, 
but since $\IN(P,T)$ does not occur in 
the plan-body, Algorithm 2 does not 
consider the literal. We do not expect 
this to be an issue in practice, however,
because such literals are accounted for 
by the event-goal's (extracted) precondition. 

The third case is where must literals 
are ``hidden'' due to the particular
variable/constant symbols chosen by the 
domain writer when encoding literals. For 
example, given the following two plan-rules 
for an event-goal that sends an email 
from $F$ to $T$, literal $\sent(T)$ 
is only a mentioned literal of 
$\sendEmail(F, T)$ according to 
Algorithm 3 (line 7 in particular), 
but a must literal of it by definition:
\[
\begin{array}{ll}
\hspace{-1.5mm}   \plane{\sendEmail(F,T): (F \neq T) <- +\addedSignature\ ; +\sent(T)},\\
\hspace{-1.5mm}   \plane{\sendEmail(F,T): (F = T) <- +\sent(F)}.\\
\end{array}
\]
  
Nonetheless, by changing $+\sent(F)$ 
to $+\sent(T)$, which then mentions the
same variable symbol as the first plan-body, 
$\sent(T)$ is identified by the 
algorithm as a must literal of $\sendEmail(F,T)$. 
In general, such ``hidden'' must literals 
can be disclosed by choosing terms with 
appropriate care.

Finally, while Algorithm 2 ``conservatively''
classifies any must literal that is may 
undone as a may literal, it could still 
be a must literal by definition. For 
example, given an event-goal $move(X,Y)$,
suppose that the following 
plan-rule is the only one relevant 
for the event-goal:
\[
\begin{array}{ll}
\plane{move(X,Y): at(X) \land \neg at(Y) <- -at(X)\ ; +at(Y)}.
\end{array}
\]
Then, by Definition \ref{def:must-summ}, 
both $\neg at(X)$ and $at(Y)$ are must 
literals of the event-goal, but only 
$at(Y)$ is its must literal according 
to Algorithm 2, because it cannot infer 
that the context condition entails 
$X \neq Y$.\footnote{Note that if the 
context condition is just $at(X)$, then,
by definition, $at(Y)$ would indeed be 
the only must literal of the event-goal, 
because it would then be possible for 
$X$ and $Y$ to have the same value, 
and for $at(Y)$ to ``undo'' $\neg 
at(X)$.}  
While the algorithm does fail to 
detect some must literals in such 
domains, this can sometimes be
averted by encoding the domain
differently.
For example, the above rule can be 
encoded as an action-rule instead,
in which case Algorithm 1 (in line 1) 
will classify $\neg at(X)$ (and $at(Y)$)
as a must literal of $move(X,Y)$, under 
the assumption that action-rules
are coherent.

%%%%%%%%%%%%%%%%%%%%%%%%%%%%%%%%%%%%%%%%%%%%%%%%%%%%%%%%%%%%%%%%%%%%%%%%%%%%%%%%%%
%%%%%%%%%%%%%%%%%%%%%%%%%%%%%%%%%%%%%%%%%%%%%%%%%%%%%%%%%%%%%%%%%%%%%%%%%%%%%%%%%%
%%%%%%%%%%%%%%%%%%%%%%%%%%%%%%%%%%%%%%%%%%%%%%%%%%%%%%%%%%%%%%%%%%%%%%%%%%%%%%%%%%
\section{AN APPLICATION TO PLANNING}
\label{sec:find-hyb-plans}

One application of the algorithms presented is to 
create abstract planning operators that may be used 
together with primitive operators and a classical 
planner in order to obtain abstract (or ``hybrid'') 
plans. While \cite{desilva:aamas:09} focuses on
algorithms for extracting an ``ideal'' abstract 
plan from an abstract plan that is supplied, 
here we give the details regarding how a first abstract 
plan may be obtained.
 
To get abstract operators $\Lambda^a$ from a
plan-library $\Pi$ and an action-library 
$\Lambda$, we take the set 
$\Delta = \Summarise(\Pi,\Lambda)$ and
create an (abstract) operator for every 
summary information tuple 
$\tuple{e,\phi,L^{mt},L^{mn}} \in \Delta$.
To this end, we take the operator's name
as $e$, appended with its arity and combined 
with any additional variables occurring in 
$\phi$; the operator's precondition as $\phi$; 
and its postcondition as the set of must 
literals $L^{mt}$.

Since mentioned literals of event-goals are not 
included in their associated abstract operators, it 
is crucial that we ascertain whether these literals 
will cause unavoidable conflicts in an abstract plan 
found. For example, consider the classical 
planning problem with initial state $p$ and goal 
state $r$, and the abstract plan $e_1 \cdot e_2$ 
consisting of two event-goals (or abstract 
operators). Suppose $e_1$ and $e_2$ have 
the following plan-rules:
\[\small
\begin{array}{ccc}
\plane{e_1:\true<-+p;+q} & 
\plane{e_1:\true<--p;+q} & 
\plane{e_2:p \land q <-+r}
\end{array}
\]

Notice that the postconditions (must literals) of 
abstract operators $e_1$ and $e_2$ are respectively 
$q$ and $r$, and that $e_1 \cdot e_2$ is a classical 
planning solution for the given planning problem. 
However, when this plan is executed, if $e_1$ is 
decomposed using its second plan-rule, this will 
cause (mentioned literal) $\neg p$ to be brought 
about, thereby invalidating the context condition 
of $e_2$ (which requires $p$).
 
To check for such cases, we present the following 
simple polynomial-time algorithm. Suppose that
$P = P_1 ; \ldots ; P_n$ 
is the program corresponding to a classical
planning solution $P'_1 \cdot\ldots\cdot P'_n$ 
for some planning problem, where each (ground) 
$P_i$ is either an action or event-goal. Then, 
we say that $P$ is \defterm{correct} relative to 
$\Delta$ if for any (ground) literal $l$ occurring 
in the precondition of any $P_i$,
the following condition holds: if $l$ is not must 
undone and it is may undone (relative to $\Delta$)
in the preceding subplan 
$P_{1} ; \ldots ; P_{i-1}$ by some mentioned literal 
$l'$ of a step $P_k$ in the subplan,\footnote{We rely
here on a slightly extended version of the definition
of may undone from before, to have the exact step 
($P_k$) and literal ($l'$) responsible for the 
``undoing''. Moreover, observe that literals $l$ 
and $l'$ are 
obtained by applying the same substitution that 
the planner applied to obtain $P'_i$ and 
$P'_k$, respectively.} then literal $l$, or its 
complement, is also must undone (relative to $\Delta$) 
in the steps $P_{k+1} ; \ldots ; P_{i-1}$.
Otherwise, $P$ is said to be \defterm{potentially 
incorrect}. Interestingly, the situation where $l$ 
is must undone in $P_1 ; \ldots ; P_{i-1}$ is not
unacceptable because it cannot invalidate the 
(possibly disjunctive) precondition of $P_i$, 
given that $P_1 ; \ldots ; P_n$ corresponds 
to a solution for some classical planning problem.
The following theorem states that, as expected, 
a correct program $P$ will always have at least 
one successful HTN decomposition.

\begin{theorem}
Let $\Pi$ be a plan-library, $\Lambda$ an action-library, 
$\Delta = \Summarise(\Pi,\Lambda)$, and $P$ the
program corresponding to a
solution for the classical planning 
problem $\tuple{\BB,\BB_g,\Lambda \cup \Lambda^a}$,
where $\BB$ and $\BB_g$ are belief bases representing
respectively initial and goal states.
Then, if $P$ is correct (relative to $\Delta$), 
there is a successful HTN 
execution $C_1 \cdot\ldots\cdot C_n$ of
$P$ such that $C_1|_{\BB} = \BB,
C_1|_{P} = P$ and $C_n|_{\BB} \models \BB_{g}$. 
\end{theorem}
\begin{myproof}
If there is no such successful execution, since
$P=P_1; \ldots ;P_n$ is a classical 
planning solution, there must be a mentioned 
literal of some $P_i$ that intuitively ``conflicts'' 
with a literal occurring in the precondition of 
some $P_j$, with $j>i$. The classical planner 
will not have taken such conflicts into account, 
but according to the definition of what 
it means for $P$ to be correct, such a mentioned
literal cannot exist.
\end{myproof}

If we find that $P$ is potentially incorrect, we 
then determine whether it is \textit{definitely 
incorrect}, i.e., whether there are conflicts that 
are unavoidable. To this end, we look for a successful 
HTN decomposition of $P$, failing which the plan 
is discarded and the process repeated with a new 
abstract plan.

\section{DISCUSSION \& FUTURE WORK}

We have presented definitions and sound algorithms 
for summarising plan hierarchies which, unlike past
work, are defined in a typical and well understood 
BDI agent-oriented programming language. By virtue 
of its syntax and semantics being inherently tied 
to HTN planning, our work straightforwardly applies 
to HTN planners such as SHOP \cite{shop2:03}. Our 
approach is closely related to \cite{clement:jair:07}, 
the main differences being that we support variables
in agent programs, and we reason about non-concurrent 
plans. While these do make a part of our approach 
incomplete, we have shown how this can sometimes 
be averted by writing domains with appropriate 
care. 
Crucially, we have handled variables ``natively'',
without grounding them on a finite set of constants. 
We concluded with one application of our 
algorithms, showing how they can be used 
together with a classical planner in order 
to obtain abstract plans. 

We expect that the summaries we compute 
will be useful in other applications that 
rely on similar information, such as coordinating 
the plans of single \cite{john-ij,John-03} 
and multiple \cite{clement:jair:07} agents, 
and particularly in improving HTN planning 
efficiency \cite{tsuneto:98}. There is also
potential for using such information as 
guidance when creating agent plans manually 
\cite{yao:16}. 

Interestingly, the application we presented
mitigates our restriction that plan-libraries
cannot be recursive, as the classical planner 
can, if necessary, repeat an event-goal in 
an abstract plan.
Nonetheless, allowing recursive plan-libraries 
is still an interesting avenue for future work. 
Another useful improvement would be to allow 
partially ordered steps in plan-bodies (i.e.,
the construct $P \parallel P'$). Given a 
plan-library $\Pi$, one potential approach 
to that end is to obtain the plan-library 
$\Pi'$ consisting of all linear extensions 
of plan-rules in $\Pi$, and then use $\Pi'$ 
as the input into Algorithm 1 (\Summarise). 
We could use existing, fast algorithms to 
generate linear extensions \cite{Pruesse:1994}, 
or consider simpler plan-rules corresponding 
to restricted classes of partially ordered 
%sets (e.g. series-parallel or bounded width 
sets \cite{Bouchitte:1989}.
Finally, it would be interesting to formally
characterise the restricted class of domains 
in which the presented algorithms are complete. 

\section {ACKNOWLEDGEMENTS}
This work was supported by Agent Oriented Software and the Australian Research Council (grant LP0882234). We thank Brian Logan for useful discussions relating to the work presented, and the anonymous reviewers for helpful feedback.
\bibliographystyle{abbrv}

%\bibliography{bib-keys-full,bib-ssardina,summ}

\begin{thebibliography}{10}

\bibitem{wayne:09}
R.~W. Alford, U.~Kuter, and D.~Nau.
\newblock Translating {HTN}s to {PDDL}: A small amount of domain knowledge can
  go a long way.
\newblock In {\em Proceedings of the International Joint Conference on
  Artificial Intelligence (IJCAI-09)}, pages 1629--1634, 2009.

\bibitem{sheila:icaps:07}
J.~A. Baier, C.~Fritz, and S.~A. McIlraith.
\newblock Exploiting procedural domain control knowledge in state-of-the-art
  planners.
\newblock In {\em Proceedings of the International Conference on Automated
  Planning and Scheduling (ICAPS-07)}, pages 26--33, 2007.

\bibitem{macroff}
A.~Botea, M.~Enzenberger, M.~M{\"u}ller, and J.~Schaeffer.
\newblock Macro-{FF}: Improving {AI} planning with automatically learned
  macro-operators.
\newblock {\em Journal of Artificial Intelligence Research (JAIR)},
  24:581--621, 2005.

\bibitem{Bouchitte:1989}
V.~Bouchitte and M.~Habib.
\newblock The calculation of invariants for ordered sets.
\newblock In I.~Rival, editor, {\em Algorithms and Order}, volume 255, pages
  231--279. Springer Netherlands, 1989.

\bibitem{ceballos2011genom}
A.~Ceballos, L.~de~Silva, M.~Herrb, F.~Ingrand, A.~Mallet, A.~Medina, and
  M.~Prieto.
\newblock Genom as a robotics framework for planetary rover surface operations.
\newblock In {\em Symposium on Advanced Space Technologies in Robotics and
  Automation (ASTRA)}, 2011.

\bibitem{clement:jair:07}
B.~J. Clement, E.~H. Durfee, and A.~C. Barrett.
\newblock Abstract reasoning for planning and coordination.
\newblock {\em Journal of Artificial Intelligence Research (JAIR)},
  28:453--515, 2007.

\bibitem{desilva:aamas:09}
L.~de~Silva, S.~Sardina, and L.~Padgham.
\newblock First {P}rinciples {P}lanning in {BDI} systems.
\newblock In {\em Proceedings of the International Joint Conference on
  Autonomous Agents and Multiagent Systems (AAMAS-09)}, pages 1105--1112, 2009.

\bibitem{erol94htn}
K.~Erol, J.~Hendler, and D.~S. Nau.
\newblock {HTN} planning: Complexity and expressivity.
\newblock In {\em Proceedings of the National Conference on Artificial
  Intelligence ({AAAI}-94)}, pages 1123--1128, 1994.

\bibitem{sheila:kr:08}
C.~Fritz, J.~A. Baier, and S.~A. McIlraith.
\newblock {C}on{G}olog, {S}in {T}rans: Compiling {C}on{G}olog into {B}asic
  {A}ction {T}heories for planning and beyond.
\newblock In {\em Proceedings of the International Conference on Principles of
  Knowledge Representation and Reasoning (KR-08)}, pages 600--610, 2008.

\bibitem{gabbay:book}
D.~M. Gabbay, C.~J. Hogger, and J.~A. Robinson, editors.
\newblock {\em Handbook of {L}ogic in {A}rtificial {I}ntelligence and {L}ogic
  {P}rogramming}.
\newblock Oxford University Press, 1994.

\bibitem{GhallabNT:04-Planning}
M.~Ghallab, D.~S. Nau, and P.~Traverso.
\newblock {\em Automated Planning: {T}heory and Practice}.
\newblock Morgan Kaufmann Publishers Inc., 2004.

\bibitem{rao:aaai:98}
S.~Kambhampati, A.~D. Mali, and B.~Srivastava.
\newblock Hybrid planning for partially hierarchical domains.
\newblock In {\em Proceedings of the National Conference on Artificial
  Intelligence (AAAI-98)}, pages 882--888, 1998.

\bibitem{lloyd:87}
J.~W. Lloyd.
\newblock {\em Foundations of Logic Programming; (2nd Extended Ed.)}.
\newblock Springer-Verlag New York, Inc., 1987.

\bibitem{Martelli:1982}
A.~Martelli and U.~Montanari.
\newblock An efficient unification algorithm.
\newblock {\em ACM Transactions on Programming Languages and Systems},
  4(2):258--282, 1982.

\bibitem{shop2:03}
D.~S. Nau, T.-C. Au, O.~Ilghami, U.~Kuter, J.~W. Murdock, D.~Wu, and F.~Yaman.
\newblock {SHOP}2: An {HTN} planning system.
\newblock {\em Journal of Artificial Intelligence Research (JAIR)},
  20:379--404, 2003.

\bibitem{plotkin}
G.~D. Plotkin.
\newblock A structural approach to operational semantics.
\newblock Technical Report DAIMI FN-19, Computer Science Department, University
  of Aarhus, Denmark, 1981.

\bibitem{Pruesse:1994}
G.~Pruesse and F.~Ruskey.
\newblock Generating linear extensions fast.
\newblock {\em SIAM Journal on Computing}, 23(2):373--386, 1994.

\bibitem{agentspeak}
A.~S. Rao.
\newblock {A}gent{S}peak({L}): {BDI} agents speak out in a logical computable
  language.
\newblock In {\em Proceedings of the European workshop on Modelling Autonomous
  Agents in a Multi-Agent World : agents breaking away (MAAMAW-96)}, pages
  42--55. Springer, 1996.

\bibitem{rao:95}
A.~S. Rao and M.~P. Georgeff.
\newblock {BDI}-agents: from theory to practice.
\newblock In {\em Proceedings of the International Conference on Multiagent
  Systems (ICMAS-95)}, pages 312--319, 1995.

\bibitem{sardina:aamas:06}
S.~Sardina, L.~de~Silva, and L.~Padgham.
\newblock Hierarchical planning in {BDI} agent programming languages: A formal
  approach.
\newblock In {\em Proceedings of the International Joint Conference on
  Autonomous Agents and Multiagent Systems (AAMAS-06)}, pages 1001--1008, 2006.

\bibitem{sardina:jaamas}
S.~Sardina and L.~Padgham.
\newblock A {BDI} agent programming language with failure handling, declarative
  goals, and planning.
\newblock {\em Autonomous Agents and Multiagent Systems}, 23(1):18--70, 2011.

\bibitem{john-ij}
J.~Thangarajah, L.~Padgham, and M.~Winikoff.
\newblock Detecting and avoiding interference between goals in intelligent
  agents.
\newblock In {\em Proceedings of the International Joint Conference on
  Artificial Intelligence (IJCAI-03)}, pages 721--726, 2003.

\bibitem{John-03}
J.~Thangarajah, L.~Padgham, and M.~Winikoff.
\newblock Detecting and exploiting positive goal interaction in intelligent
  agents.
\newblock In {\em Proceedings of the International Joint Conference on
  Autonomous Agents and Multiagent Systems (AAMAS-03)}, pages 401--408, 2003.

\bibitem{tsuneto:98}
R.~Tsuneto, J.~Hendler, and D.~Nau.
\newblock Analyzing external conditions to improve the efficiency of {HTN}
  planning.
\newblock In {\em Proceedings of the National Conference on Artificial
  Intelligence (AAAI-98)}, pages 913--920, 1998.

\bibitem{Winikoff02}
M.~Winikoff, L.~Padgham, J.~Harland, and J.~Thangarajah.
\newblock Declarative and procedural goals in intelligent agent systems.
\newblock In {\em Proceedings of the International Conference on Principles of
  Knowledge Representation and Reasoning (KR-02)}, pages 470--481, 2002.

\bibitem{yao:16}
Y.~Yao, L.~de~Silva, and B.~Logan.
\newblock Reasoning about the executability of goal-plan trees.
\newblock In {\em Engineering Multi-Agent Systems Workshop (EMAS-16)}, pages
  181--196, 2016.

\end{thebibliography}

\section{Appendix}
The lemmas that follow rely on the following definition of what
it means for a set of literals to \textit{capture} a program.
Intuitively, a set of literals captures a program if
any literal resulting from any successful execution of the program
is in the set.

\begin{definition}[Capturing a Program]\label{def:captures}
Let $P$ be a program and $L$ be a set of literals. Set $L$ \defterm{captures}
$P$ if and only if for any ground instance $P^g$ of $P$, successful \HTN\
execution $C_1 \cdot\ldots\cdot C_n$ of $P^g$, and ground literal $l$
%
%finite sequence of configurations $E$ of the form $C_1 = \config{\BB_1,
%\CA_1, P^g} \cdot\ldots\cdot C_n = \config{\BB_n, \CA_n, nil}$ (where
%$\BB_1,\BB_n$ are belief bases and $\CA_1,\CA_n$ are action sequences),
%with $C_i \transitionhtn C_{i+1}$ for every $i > 0$,
%
such that $C_1|_{\BB} \not\models l$ and $C_n|_{\BB} \models l$, it is the
case that there is a literal $l' \in L$ such that $l = l'\theta$, for
some substitution $\theta$.
\end{definition}

Observe, then, that the (full) set of mentioned literals of a program captures
the program.

\begin{lemma}
\label{thm:atomic-prog-summ-correctness}
Let $P$ be a primitive program (i.e., $P = ?\phi \mid +b \mid -b \mid act$) 
mentioned in a plan-library $\Pi$, and let $\Lambda$ be an action-library. 
Given $\Pi$ and $\Lambda$ as input for Algorithm 1, %\ref{alg:summarise},
at the end of line \ref{alg:lin:prim-start} of the algorithm, 
there exists exactly one tuple $\tuple{P, \epsilon, L^{mt}, L^{mn}} 
\in \Delta$ such that the tuple is the summary information of $P$  
and $L^{mn}$ captures $P$.
\end{lemma}
\begin{myproof}
%\noindent\textbf{Proof}
%
We consider all possible cases for $P$.
\begin{itemize}
\item Case $P = ?\phi$. Then, $post(P) = \emptyset$, and there is exactly one 
tuple $\tuple{P, \epsilon,\emptyset,\emptyset} \in \Delta$. Since no literal 
is added to the belief base upon the execution of $P$, and $\emptyset$
is a valid set of must literals (Definition \ref{def:must-summ}),
the theorem holds.

\item Case $P = +b$. 
Then, observe that there is exactly one tuple $\tuple{P, \epsilon,\{b\},\{b\}} \in 
\Delta$ on the completion of line \ref{alg:lin:prim-start}.
Next, let $b\theta$ be any ground instance of $b$. Then, 
for all belief bases $\BB$ and action sequences $\CA$,
the following two conditions hold: \textit{(i)} $\config{\BB, \CA, +b\theta} 
\transitionhtn \config{\BB' = \BB \cup \{b\theta\}, \CA, nil}$, i.e.,
there is a successful \HTN\ execution of $+b\theta$, and \textit{(ii)} $\BB' \models 
b\theta$. Therefore, $b$ is a must literal of $P$. 
%%%% CAPTURES PROOF
%Similarly, since 
%the addition of belief atom $b\theta$ to
%belief base $\BB$ is the only modification that can happen to $\BB$
%on the successful \HTN\ execution of $+b\theta$, it follows that set
%$\{b\}$ captures $P$ (Definition \ref{def:captures}), and the theorem
%holds.

\item 
Case $P = -b$. This case is proved analogously to the one above.

\item
Case $P = act$. Notice that according to the definition of an action's
postcondition, $act$ will be unified with the head of some action-rule in 
$\Lambda$ (such a unification will always be possible due to our assumption
in Section ``\nameref{sec:assum}'' that plan-libraries are written with 
appropriate care).
Then, let $\plane{act':\psi<-\Phi^+;\Phi^-} \in \Lambda$, with $act = act'\theta$,
be that action-rule. Moreover, let $\hat{\Phi}^+ = \Phi^+\theta$ and 
$\hat{\Phi}^- = \Phi^-\theta$. Finally, let the set of literals
$L_{act} = \hat{\Phi}^+ \cup \{\neg b \mid b \in \hat{\Phi}^-\}$.
Observe, then, that there exists exactly one tuple
$\tuple{P, \epsilon, L_{act}, L_{act}} \in \Delta$
on the completion of line \ref{alg:lin:prim-start}.
Let $l$ be any literal in $L_{act}$. We will now prove that $l$
is a must literal of $act$. \\

First, from the definition of an action-rule, 
free variables in $l$ will also be free in $P = act$ (a prerequisite in 
Definition \ref{def:must-summ}). Now, let $act\theta'$ be any ground 
instance of $act$, $\BB$ any belief base, and $\CA$ any action sequence.
Then, notice that if $\config{\BB, \CA, act\theta'} \transitionhtn \config{\BB' = 
(\BB \setminus \hat{\Phi}^-\theta') \cup \hat{\Phi}^+\theta', \CA \cdot act\theta', nil}$ 
holds---i.e., there is a successful \HTN\ execution of $act\theta'$---then 
$\BB' \models l\theta'$ also holds. Thus $l$ is a must literal of
$P$.
%The fact that $L_{act}$ captures $P$ follows trivially from the fact
%that for any ground literal $l'$ such that $\BB' \models l'$ and $\BB 
%\not\models l'$ hold, it is also the case that $l'$ is a ground instance
%of some literal in $L_{act}$.
\end{itemize}
\vspace{-6mm}
\end{myproof}

%The following lemma states that Algorithm \ref{alg:planbodysumm} 
%($\SummarisePlanBody$) is sound, that is, whenever it classifies a 
%literal as a must literal this is indeed the case.

\begin{lemma}\label{thm:planbodysumm-alg-correctness2}
Let $P$ be a plan-body mentioned in a plan-library $\Pi$, and let $\Lambda$
be an action-library. Let $\Delta_{P}$ be a set of tuples such that: 
\begin{enumerate}

\item for each primitive program $P'$ mentioned in $P$, there is exactly 
one tuple $\tuple{P', \epsilon, L^{mt}, L^{mn}} \in \Delta_{P}$ such that 
the tuple is the summary information of $P'$ and $L^{mn}$ captures $P'$; and 

\item for the event-goal type $e'$ of each event-goal mentioned in $P$,
there is exactly one tuple $\tuple{e', \phi, L^{mt}, L^{mn}} \in \Delta_{P}$ 
such that the tuple is the summary information of $e'$, event $e=e'\theta$,
%
%is the event type of $e$, and 
and $L^{mn}$ captures $e'$.
\end{enumerate}
Finally, let $\tuple{P', \epsilon, L^{mt}, L^{mn}} = 
\SummarisePlanBody(P,\Pi,\Lambda,\Delta_{P})$.
Then, the tuple is the summary information of $P$ and $L^{mn}$ captures $P$. 
\end{lemma}
\begin{myproof}
Consider line \ref{alg:lin:event-summ-start} of procedure $\SummarisePlanBody$.
Observe that, together with the second condition in the assumption of the 
theorem, on the completion of this line, for each event-goal program $!e$ 
mentioned in $P$, there is exactly one tuple $\tuple{!e, \phi_{e}, L_{e}^{mt}, 
L_{e}^{mn}} \in \Delta$ such that the tuple is the summary information of $!e$. 
Then, together with the first condition in the assumption of the theorem, 
we can conclude that on the completion of line \ref{alg:lin:event-summ-start}, 
for each \textit{atomic} program $P^a$ mentioned in $P$, there is exactly one 
tuple $\tuple{P^a, \phi_{P^a}, L_{P^a}^{mt}, L_{P^a}^{mn}} \in \Delta$ such 
that the tuple is the summary information of $P^a$, and $L_{P^a}^{mn}$ 
captures $P^a$.
%
%where $C_{P'}$ is a set of summary conditions; $C_{P'}$ captures $P'$; 
%and for all $\tuple{l,must} \in C_{P'}$, condition $\tuple{l,must}$ 
%is a must summary condition of $P'$. When we refer to set $\Delta$
%in this proof we are referring to the value of $\Delta$ from
%line \ref{alg:lin:event-summ-end} onwards.                    
%                   
%The proof relies on Lemmas \ref{thm:atomic-prog-summ-correctness},
%\ref{thm:eventsumm-alg-correctness} and \ref{thm:eventsumm-alg-correctness2}.
%
To prove that $\tuple{P,\epsilon,L^{mt},L^{mn}}$ is the summary
information of $P$, we will first prove that each literal in $L^{mt}$
is a must literal of $P$.
 
Let $P = P_1 ; P_2 ; \ldots ; P_n$, where each $P_i$ is an atomic
program.
Observe from line \ref{alg:lin:may-undone} of procedure $\SummarisePlanBody$
that the only literals included in the set $L_P^{mt}$ 
are the literals that are must literals $l$ of atomic programs
$P_i$ mentioned in $P$, where $l$ is not may-undone in $P_{i+1};\ldots;P_n$,
that is $\neg \MayUndone(l, P_{i+1};\ldots;P_n,$ $\Delta)$.
Let $l$ be such a literal and $P_i$ such an atomic program.
Next, we prove that $l$ is a must literal of $P$.

%Observe that, given any belief base $\BB$, sequence of actions $\CA$, and 
%ground instance $P^g$ of $P$ (where $P^g = P\theta$), a successful \HTN\
%execution $\config{\BB, \CA, P^g} \cdot\ldots\cdot \config{\BB', 
%\CA', \pnil}$, entails 
%that such a successful execution (possibly with different values for 
%$\BB,\BB',\CA,$ and $\CA'$) $E'$ of the form $\config{\BB, \CA, P_i\theta} 
%\cdot\ldots\cdot \config{\BB', \CA', \pnil}$ also exists for $P_i\theta$. 
%Moreover, since $c$ is a must summary condition of $P_i$, it is also the 
%case that $lit[c]\theta$ is asserted in $E'$ (and hence $E$) and that 
%$\BB' \models lit[c]\theta$.

%It is straightforward from Algorithms \ref{alg:may-undone} ($\MayUndone$)
%and \ref{alg:must-undone} ($\MustUndone$) that if a literal is not may-undone
%in $P_{i+1};\ldots;P_n$, then it is also not must-undone in $P_{i+1};\ldots;P_n$.
%Therefore, all we need to show is that condition $c$ is indeed a must summary
%condition of $P$ provided it is not may-undone in $P_{i+1};\ldots;P_n$.
%                   
%It is clear from lines XXX in Algorithm \ref{}, lines YYY in Algorithm
%\ref{}, and lines ZZZ in Algorithm ZZZ, that every possible literal that can
%be asserted by an execution of $P_{i+1};\ldots;P_n$ is a literal $lit[c']$
%(up to variable substitution) where $c'$ is a mentioned summary condition
%of $P_{i+1};\ldots;P_n$, unless
%                   
Let us assume the contrary, i.e., that $l$ is not a must literal of $P$.
Then, informally, it must be the case that the complement of $l$ is true
at the end of a successful \HTN\ execution of program $P_{i+1};\ldots;P_n$---in 
particular, the complement of $l$ must be true at the end of at least one
successful \HTN\ execution of some atomic program mentioned in $P_{i+1};\ldots;P_n$.
More precisely, according to Definition \ref{def:must-summ} (Must Literal), 
this means that there is an atomic program $P_j$ mentioned in 
$P_{i+1};\ldots;P_n$, a ground instance $P_j^g$ of $P_j$, and a 
successful \HTN\ execution $\config{\BB_1, \CA_1, P_j^g} 
\cdot\ldots\cdot \config{\BB_m, \CA_m, \pnil}$ of $P_j^g$ 
such that $\BB_m \models \overline{l}\theta$ for some ground 
substitution $\theta$, and $\BB_1 \not\models \overline{l}\theta$. 
%
%and \textit{(c)} $\overline{l}\theta = l'$, for some ground literal
%$l'$ and set of substitutions $\theta$.
%\begin{eqnarray}\label{prf:planbodysumm:claim1}
%
Then, by Definition \ref{def:captures} (Capturing a Program),
%Then, by the definition ``Capturing a Program'',
and using the fact that the set $L_{P_j}^{mn}$ of tuple 
$\tuple{P_j,\phi_j,L_{P_j}^{mt},L_{P_j}^{mn}} \in \Delta$ 
captures $P_j$ (by the assumption of the theorem),
% 
%We know from the assumption of the theorem that $L_{P_j}^{mn}$ captures 
%$P_j$. Then, according to Definition \ref{def:captures} (Capturing a Program), 
% 
it must also be the case that there is a literal $l' \in L_{P_j}^{mn}$ 
such that $l'\theta' = \overline{l}\theta$ for some $\theta'$. 
Next, we show that this cannot be the case.
%
%for
%some set of substitutions $\hat{\theta}$. Then, using \textit{(c)}
%above, we can conclude that $\overline{l}\theta = \hat{l}\hat{\theta}$.
%
%\begin{eqnarray}\label{prf:planbodysumm:claimtwo} \end{eqnarray}
%suppose, then, that $\models (l\theta' = \hat{l}\theta')$ holds
%for some set of substitutions $\theta$.
 
Observe that $\neg \MayUndone(l, P_{i+1};\ldots;P_n,\Delta)$ 
holds according to line \ref{alg:lin:may-undone} in procedure 
$\SummarisePlanBody$. Then, from the definition of $\MayUndone$,
%
%(Section \ref{sec:algos}) that there are no sets 
%of substitutions $\theta',\theta''$ 
%such that $\overline{l}\theta' = \hat{l}\theta''$, 
%
there is no substitution $\theta''$ such that $l\theta'' 
= \overline{l^r}\theta''$ holds, or equivalently, such that 
$\overline{l}\theta'' = l^r\theta''$ holds, where $l^r$ is 
obtained from $l'$ by renaming its variables to those that
do not occur in $l$. Observe that if there is no such 
$\theta''$, then there will also not exist two 
substitutions $\theta_1$ and $\theta_2$ such that 
$\overline{l}\theta_1 = l^r\theta_2$ holds (because
otherwise we can always combine them to form $\theta'' =
\theta_1 \cup \theta_2$). However, recall from before that 
$\overline{l}\theta =  l'\theta'$ also holds. This means
that---since $l\theta$ is ground---we can always rename 
variables in the pair $l',\theta'$ to obtain 
the pair $l^r,\theta^r$ with $\overline{l}\theta = 
l^r\theta^r$, which means that there \textit{are} two
substitutions $\theta_1,\theta_2$ such that 
$\overline{l}\theta_1 = l^r\theta_2$. This 
contradicts our assumption; therefore, literal 
$l$ is indeed a must literal of $P$.
%
%Finally, since we can always rename
%variables in $\theta''_2$ and $l^r$ so that $l^r = l'$,
%it cannot be the case that $\overline{l}\theta =  l'\theta'$
%
%contradicts Equation \ref{prf:planbodysumm:claimtwo}

Next, we prove that the set of mentioned literals $L^{mn}$ 
of program $P$ captures $P$. First, observe from line 
\ref{alg:lin:must-undone} of procedure $\SummarisePlanBody$ 
that all mentioned literals in the summary information of all 
atomic programs of $P$ are added to the set of mentioned literals 
of $P$, unless the literal is must undone. Therefore, all we 
need to show is that any must or mentioned literal of an atomic 
program occurring in $P$ that is not included in $L_{P}^{mn}$
(line \ref{alg:lin:must-undone}) is not needed for $L_{P}^{mn}$
to capture $P$.
Then, let $P_i$ be an atomic program mentioned in $P$, with $\tuple{P_i, 
\phi_i, L_{P_i}^{mt}, L_{P_i}^{mn}} \in \Delta$, such that a must
or mentioned literal $l$ of $P_i$ is not added to the set created
in line \ref{alg:lin:must-undone} of the procedure, that is, 
$\MustUndone(l,P_{i+1};\ldots;P_n,\Delta)$ holds. According to the 
definition of $\MustUndone$, this means that $\overline{l} = l'$ holds, 
where $l' \in L_{P'}^{mt}$ is a must literal of some atomic program 
$P'$ mentioned in $P_{i+1};\ldots;P_n$, with $\tuple{P', \phi_{P'}, 
L_{P'}^{mt}, L_{P'}^{mn}} \in \Delta$. Next, let
$P\theta$ be any ground instance of $P$.
%
%let $\BB$ be some belief base, and let $\CA$ be some sequence of 
%actions.
%
Suppose that a successful \HTN\ execution $\config{\BB, \CA, P_i\theta} 
\cdot\ldots\cdot \config{\BB_j, \CA_j, \pnil}$ of $P_i\theta$ exists,
such that $\BB_j \models l\theta$ holds.
%
%Observe then, from Definition \ref{def:must-summ} (Must Literal),
%that given any belief base $\BB$ and action sequence $\CA$, if a 
%successful \HTN\ execution $\config{\BB, \CA, P_i\theta} \cdot
%\ldots\cdot \config{\BB_j, \CA_j, \pnil}$ exists, then $\BB_j 
%\models l\theta$ will hold.
%
Then, since $l'$ is a must literal of $P'$, it is the case that 
$\BB'_k \models l'\theta$ also holds for any successful
\HTN\ execution $\config{\BB', \CA', P'\theta} \cdot\ldots\cdot 
\config{\BB'_k, \CA'_k, \pnil}$ of $P'\theta$. However, since 
$\overline{l}\theta = l'\theta$, literal $l\theta$ is
guaranteed to be removed from the belief base by $P'\theta$
during any successful \HTN\ execution of $P\theta$.
Therefore, a set of literals that captures $P$ does not need to
include mentioned literal $l$ of $P_i$.
(Note, however, that it is still possible that the same literal $l''$ 
from the set of must or mentioned literals of some \textit{other}
atomic program $P''$ occurring after $P'$ in $P$ will be present
in the set of literals that captures $P$, provided $l''$ is not
must undone.)
\end{myproof}

%Next, we move on to Algorithm \ref{alg:eventsumm} ($\SummariseEvent$). The 
%following two lemmas state that this algorithm is sound, that is, whenever 
%it classifies a literal as a must literal this is indeed the case, and that 
%the algorithm correctly computes summary preconditions of event-goals.
%the formula that it
%computes as the summary precondition 
%that it computes for an event-goal is indeed a summary precondition of 
%the event-goal.

\begin{lemma}\label{thm:eventsumm-alg-correctness2}
Let $e$ be the event type of some event-goal mentioned in a plan-library 
$\Pi$. Let $\Delta$ be any set such that for each plan-rule $\plane{e':\psi<-P} 
\in \Pi$ with $e = e'\theta$, there is exactly one tuple $\tuple{P, \epsilon, 
L^{mt}, L^{mn}} \in \Delta$ such that the tuple is the summary information of 
$P$ and $L^{mn}$ captures $P$. 
Finally, let tuple $\tuple{e', \phi, L^{mt}, L^{mn}} = \SummariseEvent(e,\Pi,\Delta)$.
Then, $L^{mt}$ is a set of must literals of $e$  
%
%it is the case that $e = e'$, $L^{mt}$ is a set of must literals of $e$. 
%
and $L^{mn}$ captures $e$.  
\end{lemma}
\begin{myproof}
%
%First, we will prove that $L^{mt}$ is a set of must literals of $e$.
%Let the set of plan-rules $R = \{\plane{e'\theta:\psi\theta<-P\theta} 
%\mid \plane{e':\psi<-P} \in \Pi, e = e'\theta,$ $\theta$ is a renaming 
%substitution for $\plane{e':\psi<-P}\}$.
%
Let literal $l \in L^{mt}$ and let $!e\theta$ be any ground instance of 
$!e$. If there is no successful \HTN\ execution of $!e\theta$ (relative 
to $\Pi$) then the theorem holds.  Otherwise, a successful \HTN\ execution 
of $!e\theta$ does exist. Then, by the antecedent of the \textit{Sel} 
transition rule, let $\plane{e':\psi'<-P'} \in \Pi$ be any plan-rule 
such that $e=e'\theta_r$, where $\theta_r$ is a variable renaming 
substitution for the plan-rule. Moreover, let us take the renamed plan-rule 
$\plane{e:\psi<-P} = \plane{e'\theta_r:\psi'\theta_r<-P'\theta_r}$. 
Then, by the \textit{Event}, \textit{Sel} and other transition rules, 
the following sequence of transitions must exist for some $\BB$, $\CA$, and set $D$:  
 
{\small
\[
\begin{array}{lll}
\textit{(1)}\ \config{\BB,\CA,!e\theta} 
	&      \transitionhtn       
	& \config{\BB,\CA,\pguardaltl{\{\psi\theta:P\theta,\ldots\}}},\\
\textit{(2)}\ \config{\BB,\CA,\pguardaltl{\{\psi\theta:P\theta,\ldots\}}}          
	& 	\transitionhtn
	& 	\config{\BB,\CA,P\theta\theta' \rhd \pguardaltl{D}}, \text{and}\\ 
\textit{(3)}\ \config{\BB,\CA,P\theta\theta' \rhd \pguardaltl{D}}
	&	\transitionhtnstar 
	&	\config{\BB',\CA',\pnil}.	
\end{array}
\]
}
 
%$\plane{e:\psi<-P} \in R$ such that $\config{\BB,\CA,!e\theta} 
%\transitionhtn \config{\BB,\CA,\pguardaltl{\{\psi\theta:P\theta,
%\ldots\}}} \transitionhtn \config{\BB,\CA,P\theta\theta' \rhd 
%\pguardaltl{\Delta}} \transitionhtnstar \config{\BB',\CA',\pnil}$
%holds (up to variable renaming of plan-library $\Pi$).
%
%
%Moreover, from line \ref{fig:lin:post-sum-end2} of procedure
%
%Observe from line \ref{alg:lin:smust1} of procedure $\SummariseEvent$
%that since $l \in L^{mt}$ then by the end of the loop, $l \in S'$ for
%some $S' \in S$. Moreover, 

Therefore, all we need to show is that $\BB' \models l\theta$. 
To this end, since we know that $l \in L^{mt}$, 
it is not difficult to see from lines \ref{fig:lin:post-sum-end2}, 
\ref{fig:lin:post-sum-end} and \ref{alg:lin:smust1} of procedure 
$\SummariseEvent$ that $l$ is a must literal of $P$ (up to the
renaming of variables that do not occur in $e$).
%
%(line \ref{alg:lin:smust1}).
%
By the definition of a must literal (Definition \ref{def:must-summ}),
the antecedent of the \textit{Sel} transition rule, and by virtue of
the fact that \textit{(3)} holds above, it follows that
$\BB' \models l\theta\theta'$ holds; therefore, $\BB' \models l\theta$
also holds by the definition of the composition of substitutions and
by the fact that $l\theta$ is ground (since $e\theta$ is ground).
\end{myproof}

\begin{lemma}\label{thm:eventsumm-alg-correctness3}
Let $e$ be the event type of some event-goal mentioned in a 
plan-library $\Pi$, and let $\tuple{e', \phi, L^{mt}, L^{mn}}$ 
$= \SummariseEvent(e,\Pi,\Delta)$, for some $\Delta$.
Then, $\phi$ is the precondition of $e$.
\end{lemma}
\begin{myproof}
Suppose we create a set of pairs $SP$ as follows. First, we set $SP = \emptyset$.
Next, for each plan-rule $\plane{e':\psi<-P} \in \Pi$ such that $e=e'\theta$, we 
set $SP$ to $SP \cup \{(\psi\theta,P\theta)\}$, where $\theta$ is a renaming substitution 
that renames all variables occurring in the plan-rule (except those occuring in 
$e'$, which may or may not be renamed) to variables not occurring anywhere else.
%
%Otherwise, according to line \ref{alg:lin:disj} of procedure $\SummariseEvent$, 
%formula $\phi$ is of the form
%$\phi = \bigvee_{\plane{e':\psi<-P} \in \Pi\ \land\ e=e'\theta} \psi\theta$,
%where each such $\theta$ renames all variables in the corresponding 
%plan-rule---that do not occur in the event-goal at the head of the 
%rule---to variables that do not occur anywhere else.
%
%
%Let $R = \{\plane{e'\theta:\psi\theta<-P\theta} \mid \plane{e':\psi<-P} 
%\in \Pi, e = e'\theta,\ \theta$ is a renaming substitution for
%$\plane{e':\psi<-P}\}$. Then, $\phi' = \psi_1 \lor \ldots \lor \psi_n$
%according to procedure $\SummariseEvent$, where $R = \{\plane{e_1:
%\psi_1<-P_1},\ldots,\plane{e_n:\psi_n<-P_n}\}$, and $\phi'$ is some
%variable renaming of $\phi$. 
%
Then, observe that either $\phi = \bigvee_{(\psi,P) \in SP} \psi$, or 
$\phi = \false$ if $SP = \emptyset$. 
We shall now show that $\phi$ is the precondition of $e$.
Let $!e\theta$ be any ground instance of $!e$, $\BB$ any
belief base, and $\CA$ any action sequence.
If $SP = \emptyset$, then $\phi = \false$ and $\BB \models \phi\theta$ 
does not hold and the theorem holds.
Observe from Definition \ref{def:summ-prec} (Precondition) that there
are two cases to consider.\\

\noindent [Case $\Rightarrow$]
Suppose $SP = \emptyset$. Then $\phi = \false$; therefore,
$\BB \not\models \phi\theta$, and the theorem holds trivially.

Suppose that $\BB \models \phi\theta$ holds. Then, $\BB \models 
\psi\theta\theta'$ must also hold for some disjunct $\psi$ of 
$\phi$. Let $P$ be a plan-body such that $(\psi,P) \in SP$.
Then, since $\BB \models \psi\theta\theta'$, we know from rules 
\textit{Event} and \textit{Sel} that the following transitions
are possible (up to variable renaming) for some set $D$: 
%$\config{\BB,\CA,!e\theta} \transitionhtn \config{\BB,\CA,
%\pguardaltl{\{\psi\theta:P\theta,\ldots\}}} \transitionhtn 
%\config{\BB,\CA,P\theta\theta' \rhd \pguardaltl{\Delta}}$.

{\small
\[
\begin{array}{lll}
\config{\BB,\CA,!e\theta} 
        &      \transitionhtn       
        & \config{\BB,\CA,\pguardaltl{\{\psi\theta:P\theta,\ldots\}}}, \text{and}\\
\config{\BB,\CA,\pguardaltl{\{\psi\theta:P\theta,\ldots\}}}          
        &       \transitionhtn
        &       \config{\BB,\CA,P\theta\theta' \rhd \pguardaltl{D}}.
\end{array}
\]
}
 
Finally, by our assumption in Definition \ref{def:safeLib}
%
%our assumption in Section \ref{sec:prelim} 
%which states that $\plane{e:\psi<-P}$ is \textit{safe}, 
%
we know that there is a successful
\HTN\ execution $C_1 \cdot\ldots\cdot C_n$ of $P\theta\theta'$ such that
$C_1|_{\BB} = \BB$. Therefore, it follows that there is also a successful
\HTN\ execution $C'_1 \cdot\ldots\cdot C'_m$ of $!e\theta$ such that
$C'_1|_{\BB} = \BB$.\\

\noindent [Case $\Leftarrow$] 
Suppose $SP = \emptyset$. Then, observe that $\config{\BB,\CA,!e\theta} \not\transitionhtn$ 
(for any $\CA$).
Therefore, there is no successful \HTN\ execution $C_1 \cdot\ldots\cdot C_m$ of $!e\theta$ such 
that $C_1|_{\BB} = \BB$, and the theorem holds trivially. 

Suppose that there does exist such a successful execution.
%Suppose that there is a successful \HTN\ execution 
%$C_1 \cdot\ldots\cdot C_m$ of $!e\theta$ such that $C_1|_{\BB} 
%= \BB$. 
Then, by the \textit{Event}, \textit{Sel} and other rules,
there must exist a pair $(\psi,P) \in SP$ such that $\BB \models 
\psi\theta\theta'$ holds and the following transitions are 
possible (up to variable renaming) for some $D$:

{\small
\[
\begin{array}{lll}
\config{\BB,\CA,!e\theta} 
        &      \transitionhtn       
        & \config{\BB,\CA,\pguardaltl{\{\psi\theta:P\theta,\ldots\}}},\\
\config{\BB,\CA,\pguardaltl{\{\psi\theta:P\theta,\ldots\}}}          
        &       \transitionhtn
        &       \config{\BB,\CA,P\theta\theta' \rhd \pguardaltl{D}}, \text{and}\\
\config{\BB,\CA,P\theta\theta' \rhd \pguardaltl{D}}
        &       \transitionhtnstar 
        &       \config{\BB',\CA',\pnil}.       
\end{array}
\]
}

%Finally, by the antecedent of the \textit{Sel} rule it must also be the 
%case that $\BB \models \psi\theta\theta'$ holds. 
%
Since $\psi$ is a disjunct of $\phi$, it follows that $\BB \models \phi\theta\theta'$ 
also holds. Therefore, $\BB \models \phi\theta$ holds (by the definition of the 
composition of substitutions).
\end{myproof}

\end{document}